\documentclass[10pt,journal]{IEEEtran}

\usepackage{multicol}
\usepackage{multirow}
\usepackage{amsfonts} 
\usepackage{todonotes}
\usepackage{bigstrut}
\usepackage{amssymb}
\usepackage{booktabs}
\usepackage{mathtools}
\usepackage{tabularx}
\usepackage{graphicx}
\usepackage{adjustbox}
\ifCLASSINFOpdf
\else
\fi
\begin{document}

%\title{Adapt Vision Transformer for Generalized Face Anti-Spoofing with Token Difference Aggregation}
\title{S-Adapter: Generalizing Vision Transformer for Face Anti-Spoofing with Statistical Tokens}

\author{Rizhao~Cai,
    Zitong~Yu,~\IEEEmembership{Senior Member,~IEEE},
    Chenqi~Kong,
    Haoliang~Li~\IEEEmembership{Member,~IEEE},
    Changsheng~Chen,~\IEEEmembership{Senior Member,~IEEE},
    Yongjian~Hu,
    and~Alex~C.~Kot,~\IEEEmembership{Life Fellow,~IEEE}

\thanks{Manuscript received September 2023, revised April 2024. This work was done at Rapid-Rich Object Search (ROSE) Lab, School of Electrical and Electronic Engineering (EEE), Nanyang Technological University (NTU). This research is supported in part by the NTU-PKU Joint Research Institute (a collaboration between the NTU and Peking University that is sponsored by a donation from the Ng Teng Fong Charitable Foundation), the Science and Technology Foundation of Guangzhou Huangpu Development District under Grant 2022GH15. This work is also partially supported by National Natural Science Foundation of China under Grant 62371301\&62306061, and partially supported by Guangdong Basic and Applied Basic Research Foundation (Grant No. 2023A1515140037) and Chow Sang Sang Group Research Fund under grant DON-RMG 9229161. (Corresponding author: Zitong Yu.)
}

\thanks{Rizhao Cai, Chenqi Kong, and Alex Kot are with the ROSE Lab, School of EEE, Nanyang Technological University (\{rzcai,chenqi.kong,eackot\}@ntu.edu.sg).}% <-this % stops a space
\thanks{Zitong Yu is with the School of Computing and Information Technology, Great Bay University, China (yuzitong@gbu.edu.cn).}
\thanks{Haoliang Li is with the Department of Electrical Engineering, City University of Hong Kong (haoliang.li@cityu.edu.hk).}% <-this % stops a space
\thanks{Changsheng Chen is with the Shenzhen University, China (cschen@szu.edu.cn)}
\thanks{Yongjian Hu is with the School of Electronic and Information Engineering, South China University of Technology, Guangzhou, China, and with China-Singapore International Joint Research Institute (email: eeyjhu@scut.edu.cn). }

}

\maketitle


\begin{abstract}
Face Anti-Spoofing (FAS) aims to detect malicious attempts to invade a face recognition system by presenting spoofed faces. State-of-the-art FAS techniques predominantly rely on deep learning models but their cross-domain generalization capabilities are often hindered by the domain shift problem, which arises due to different distributions between training and testing data. In this study, we develop a generalized FAS method under the Efficient Parameter Transfer Learning (EPTL) paradigm, where we adapt the pre-trained Vision Transformer models for the FAS task. During training, the adapter modules are inserted into the pre-trained ViT model, and the adapters are updated while other pre-trained parameters remain fixed. We find the limitations of previous vanilla adapters in that they are based on linear layers, which lack a spoofing-aware inductive bias and thus restrict the cross-domain generalization. To address this limitation and achieve cross-domain generalized FAS, we propose a novel Statistical Adapter (S-Adapter) that gathers local discriminative and statistical information from localized token histograms. To further improve the generalization of the statistical tokens, we propose a novel Token Style Regularization (TSR), which aims to reduce domain style variance by regularizing Gram matrices extracted from tokens across different domains. Our experimental results demonstrate that our proposed S-Adapter and TSR provide significant benefits in both zero-shot and few-shot cross-domain testing, outperforming state-of-the-art methods on several benchmark tests. We will release the source code upon acceptance.
\end{abstract}
%\keywords{face anti-spoofing, domain generalization, EPTL, ViT}

%\ccsdesc[500]{Computer systems organization~Embedded systems}

\begin{figure}
    \centering
    \includegraphics[width=\linewidth]{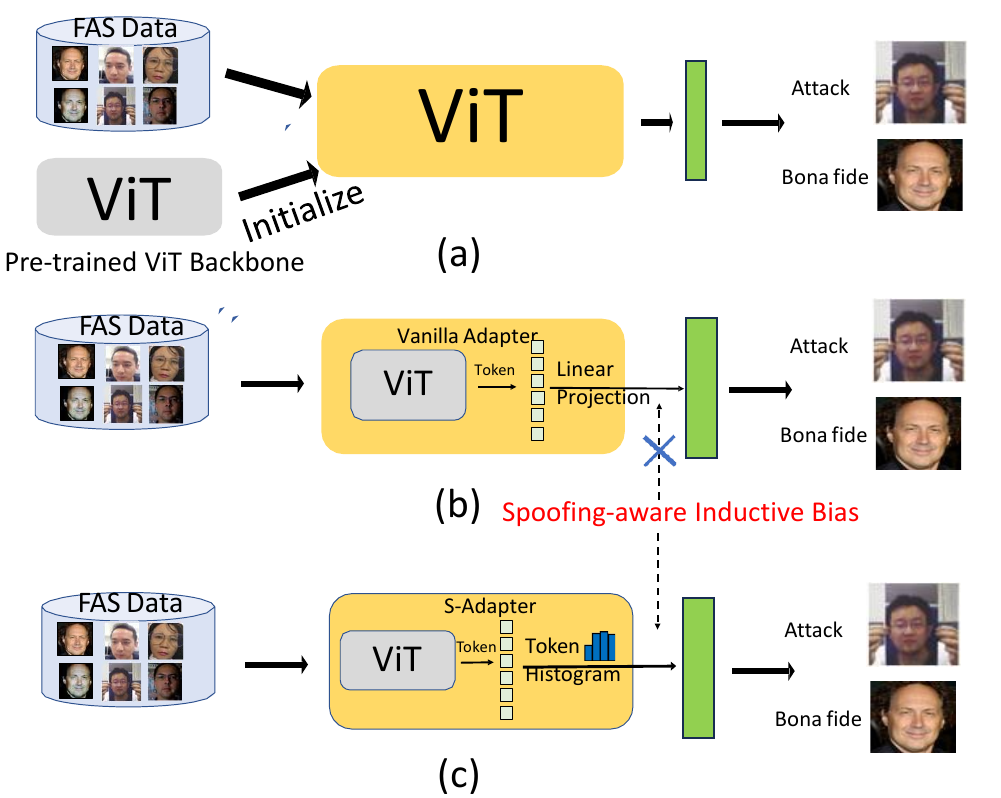}
    %\vspace{-8mm}
    \caption{(a) In the traditional transfer learning paradigm of training a ViT model for the Face Anti-Spoofing task, a pre-trained Vision Transformer (ViT) model is used for initialization, which can utilize the knowledge from the pre-training dataset. Usually, the entire or a large proportion of the model parameters are fine-tuned.  (b) In the cutting-edge Efficient Parameter Transfer Learning paradigm, adapter modules are integrated into a pre-trained ViT model. Throughout the training process, only the adapter module parameters are updated and the pre-trained parameters are fixed. Previous vanilla adapters, which are based on linear layers, lack the task-aware inductive bias \cite{chen2023visionadapter}, thereby limiting the utilization of pre-trained models. (c) Our proposed S-Adapter addresses this limitation by extracting localized token histograms to extract statistical information, enabling more efficient fine-tuning of the pre-trained ViT model for cross-domain generalized face anti-spoofing. }\label{fig:teaser}
    %\Description{figure description}
\end{figure}

\maketitle

\section{Introduction}
% Introduce face anti-spoofing
Face recognition is a convenient biometric technique and people can use their faces for identity authentication. However, face recognition systems are threatened by face presentation attacks, \textit{a.k.a}, and face spoofing attacks. These attacks involve presenting spoofing examples of human faces to cameras, such as printed photos, digital displays, and 3D masks. Face Anti-Spoofing (FAS) \cite{yu2022deep}, Face Presentation Attack Detection (Face PAD) or Face Liveness Detection, is a crucial technology that aims to enhance the security of face recognition (FR) systems by detecting malicious spoofing attacks. 

% Introduce deep learning methods and domain shift problem
To safeguard FR systems against such malicious spoofing attacks, various techniques have been extensively researched and developed \cite{yu2022deep, kong2022digital,kong2022beyond,kong2022_face_attack}. Traditional methods are mainly based on handcraft features and Support Vector Machines \cite{FAS-ColorTexture-TIFS-2016, HOG}. Given the limited representation capability of the handcrafted features, traditional methods cannot meet the security requirement of a FAS system. In recent years, deep neural networks have been increasingly incorporated into data-driven face anti-spoofing (FAS) methods to extract learnable features, surpassing traditional methods \cite{yu2022deep}. However, the deployment of these models is facing the domain shift problem, which arises from differences in data distribution between the source training data and the target testing data, caused by various data collection conditions, such as illumination, cameras, or attack mediums \cite{FAS-3DCNN-TIFS-2018}. The resulting overfitting of the training distribution can lead to poor performance on the target testing data, hindering the effective detection of spoofing attacks. To tackle the challenge in cross-domain testing, previous research has explored various techniques, including but not limited to reinforcement learning \cite{CAI-2020-DRL}, adversarial learning \cite{SSDG-CVPR-2020, SSAN}, meta-learning \cite{RFMetaFAS-AAAI-2020, MetaTeacher-TPAMI-2021, metapattern, NASFAS-TPAMI-2020}, disentanglement learning \cite{wang2020cross,liu2020disentangling}, and casual intervention \cite{liu2022causal,zheng2023learning}. Despite the progress achieved, cross-domain generalization performance remains unsatisfactory due to the critical challenge posed by the domain shift problem, and further research effort is still needed.

% Introduce previous works of ViT and fine-tuning
Recently, the cutting-edge Vision Transformer (ViT) models have achieved striking performance with the self-attention mechanism for computer vision tasks \cite{ViT, liu2021swin}. Inspired by the success of ViT, the FAS researchers have been exploring the use of ViT to address the face anti-spoofing problem \cite{george2021effectiveness, eccv2022vit}. While training a ViT model to the FAS task from scratch requires a large amount of data to achieve generalized performance, the model weights of ImageNet pre-trained ViT are easily available from open-source model zoos and can be used for model initialization for training a ViT model on FAS data \cite{george2021effectiveness}, as shown in Fig.~\ref{fig:teaser}. \textcolor{black}{However, previous works utilize the pre-trained model by fine-tuning either the partial or the whole model weights of the ViT backbone. Such utilization is straightforward but inefficient}. Recent research on Efficient Parameter Transfer Learning (EPTL) has shown a more efficient way of utilizing pre-trained ViT models for the FAS problem. Huang \textit{et al.} \cite{eccv2022vit} utilized multi-stream linear adapters to adapt ViT efficiently and achieve promising generalization performance in the few-shot cross-domain scenario. However, in the zero-shot scenario, the unseen target domain testing scenario \cite{MADDG-CVPR-2019}, the ViT's generalization performance is still inferior to the previous state-of-the-art on the four-dataset benchmark \cite{MADDG-CVPR-2019}. We identify the limitation of using vanilla adapters based on linear layers. Linear layers lack image-aware inductive bias, such as locality, and are thus ineffective in extracting local information \cite{convpass}. Since the FAS data are visual images and the local information is crucial for the classification \cite{CAI-2020-DRL}, the linear-layer-base adapter fails to capture discriminative local information to efficiently adapt ViT for FAS. Moreover, the feature/token embeddings used for FAS classification are sensitive to the imaging process, such as variations in camera modules and illuminations. Such variations between the source training and target testing data cause the domain shift and lead to models' poor domain generalization performance \cite{FAS-3DCNN-TIFS-2018}. However, it is non-trivial to learn domain-invariant information by simply using linear layers. Nevertheless, ViT with EPTL leads to a promising direction for future research in the field of FAS and deserves further development.

Motivated by the above discussion, we propose to design a more advanced adapter, named S-Adapter for the face anti-spoofing problem, to efficiently fine-tune a pre-trained ViT for cross-domain generalized FAS. As illustrated in Fig.~\ref{fig:inspiration}, our S-Adapter is motivated by traditional texture analysis methods, which collect histogram features from handcrafted feature maps, such as local binary pattern maps \cite{FAS-ColorTexture-TIFS-2016, deep-forest-lbp}, to alleviate the negative impact from varying environments, such as lightings. Our S-Adapter first extracts learnable discriminative token maps. Then token histograms are extracted, which provide statistical information and improve the robustness against variations in the environment. \textcolor{black}{Furthermore, the statistical information can benefit the model but its effectiveness is still hindered by the style variance between different domains. To reduce the style variance, we propose Token Style Regularization (TSR). The proposed TSR extracts style components based on the gram matrix, and regularizes the style variance of real faces from different domains to be minimized. As such, the statistical information with less style variance would be more generalized for the cross-domain FAS.} We conduct extensive experiments to show that our proposed method surpasses the vanilla adapter by a clear margin and achieves state-of-the-art performance on existing cross-domain face anti-spoofing benchmarks. The contributions of our work can be summarized as follows:
%%\vspace{-1mm}
\begin{itemize}
\item \textcolor{black}{We propose a novel S-Adapter to efficiently adapt pre-trained ViT models to achieve generalized face anti-spoofing by extracting statistical information via token histograms}; 
\item We propose a new Token Style Regularization (TSR), which reduces the style variances across different domains to improve the generalization of statistical token histograms.;
\item The ViT model integrated with our proposed S-Adapter and TSR can achieve state-of-the-art generalization performance on existing face anti-spoofing benchmarks, including zero/few-shot cross-domain generalization and unseen attack detection.
\end{itemize}

\section{Related works}
\subsection{Face Anti-Spoofing.}
\subsubsection{Traditional FAS Methods} 
Traditional face anti-spoofing (FAS) methods rely on handcrafted image descriptors to extract features for classification, such as Local Binary Patterns (LBP) \cite{LBP-FAS-BIOSIG-2012, FAS-ColorTexture-TIFS-2016,  deep-forest-lbp}, Histogram of Gradient \cite{HOG}, Difference of Gaussian (DoG) \cite{DoG-ECCV-2010}, and image quality features \cite{FAS-IDA-TIFS-2015, IQA-ICPR-2014, LI-QUALITY}. These pioneering methods' performance is limited by the representation capability of handcraft features, and even the intra-domain performance is not satisfactory.

\subsubsection{Deep Learning FAS Methods.} Recently, numerous FAS methods based on deep neural networks have been proposed to exploit their powerful representation learning capabilities \cite{yu2022deep, kong2022digital}. For example, reinforcement learning has been proposed to mine local and global features for FAS \cite{CAI-2020-DRL}. Pixel-wise supervision has been studied to show more advanced performance than binary supervision \cite{FAS-Auxiliary-CVPR-2018, Ternary-TIFS-2018, deeppixel--ICB-2019, yu2021revisiting}. However, models trained solely on RGB images often suffer from overfitting and poor generalization performance when there are domain shifts between training and testing data \cite{FAS-3DCNN-TIFS-2018}. Besides, hybrid methods, which combine handcraft features and deep learning have also been proposed \cite{TIFS-2019-MotionBlur,CNN-LBPTOP-2017, rehman2019perturbing, Pinto,FAS-MSR-TIFS-2019}. While the above methods have achieved saturated performance in the intra-domain evaluation, more evaluation scenarios are raised and studied, such as domain generalization scenario \cite{MADDG-CVPR-2019,MetaTeacher-TPAMI-2021,metapattern,cai2023rehearsal,lin2024suppress}, unsupervised domain adaptation scenario \cite{wang2020cross, liuYuchen2022source}, unseen attack detection scenario \cite{unseen_attack_FAS}, and so on. 

\subsubsection{FAS under Different Scenarios.} The domain generalization (DG) scenario in FAS aims to learn a model with source data from one or more domains and can achieve generalized performance on unseen target data domains without using target domain data \cite{MADDG-CVPR-2019, metapattern}. Usually, the target domain data is unseen from the training, meaning the different data distributions between training and testing. This scenario is also referred to as unseen domain generalization or zero-shot cross-domain generalization. In this scenario,  methods are expected to learn domain-invariant feature representations, and thus various techniques have been proposed to tackle domain generalization challenges in FAS, such as casual intervention \cite{liu2022causal}, disentangled representation learning \cite{liu2020disentangling,zhang2020face_dis, wu2021dual,yan2022domain}, and meta-learning \cite{NASFAS-TPAMI-2020,MetaTeacher-TPAMI-2021,metapattern,RFMetaFAS-AAAI-2020}, adversarial learning \cite{MADDG-CVPR-2019,SSDG-CVPR-2020,SSAN}, and contrastive learning \cite{HIFIMASK,SSDG-CVPR-2020,george2020learning}. The domain generalization scenario is a crucial challenge since domain shift would deter an FAS model from being deployed to practical environments. Meanwhile, the unsupervised domain adaptation (UDA) for face anti-spoofing is to utilize target domain data to adapt a model pre-trained on the source domain data but without the labels of attack and bona fide examples \cite{FAS-UnsupervisedDA-TIFS-2018,liuYuchen2022source,wang2020cross}. Since the accessibility of labels is not often a problem, the few-shot cross-domain face anti-spoofing is studied, to utilize a few labeled target domain data (\textit{e.g.} 5-shot, 10 examples) during the training to achieve great generalization performance in the target domain \cite{qin2020learning,eccv2022vit}. Likewise, one-class adaptation is also studied, where only real face examples are available \cite{li2022one-class,qin2020oneclass,george2020learning}. Moreover, other than common replay, print, and 3D mask attacks \cite{jia20203d}, more attack types appear such as makeup attacks, partial attacks, and obfuscation attacks. To evaluate a model's performance against unseen attack types, the unseen attack detection scenario has also been proposed and studied \cite{LIZHI,SiW-M-2019,unseen_attack_FAS,FAS-MultichannelCNN-TIFS-2020}. In this work, we extensively evaluate our proposed method in zero-shot cross-domain (DG) and few-shot domain generalization scenarios, as well as the unseen attack detection scenario.

\begin{figure}[t]
    \centering
    \includegraphics[width=\linewidth]{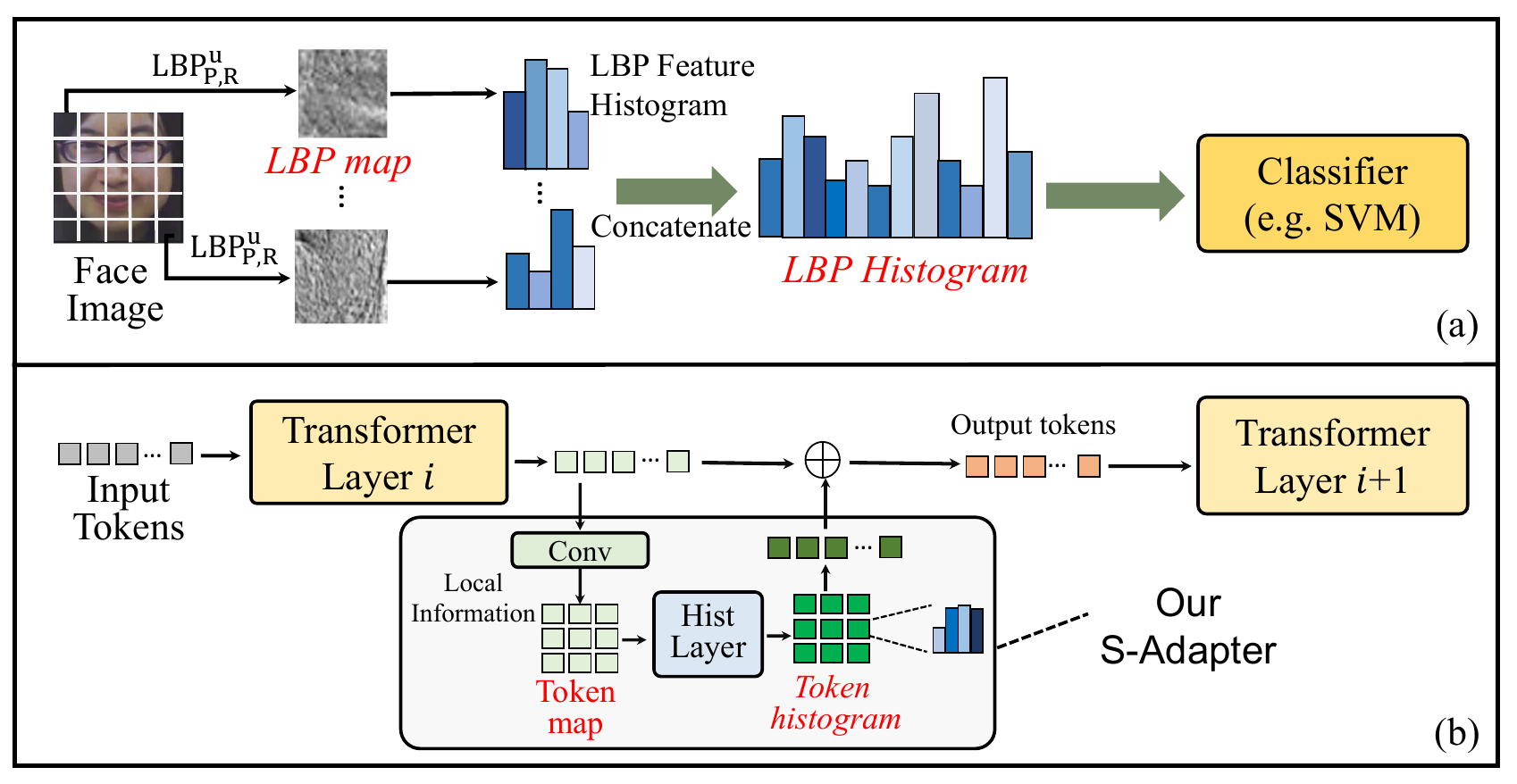}
     %\vspace{-8mm}
    \caption{(a) The process of traditional texture analysis method for face anti-spoofing: handcraft features (LBP) are first extracted, which are often sensitive to illumination changes. Then, the histogram features are extracted as final representations for the classifier, which is more robust to lighting changes. (b) Our adapter extracts local information from spatial tokens and extracts token histogram, which is inspired by (a), for improving cross-domain performance. } \label{fig:inspiration}
    \vspace{-5mm}
\end{figure}

\subsection{Efficient Parameter Transfer Learning for ViT}
The Vision Transformer (ViT) has a key component called Multi-Head Self-Attention (MHSA). In MHSA, the input token $X$ is first transformed into the query ($\mathcal{Q}$), key ($\mathcal{K}$), and value ($\mathcal{V}$): $\mathcal{Q}=X\mathcal{W}^\mathcal{Q}$, $\mathcal{K}=X\mathcal{W}^\mathcal{K}$, and $\mathcal{V}=X\mathcal{W}^\mathcal{V}$, where $\mathcal{W}^\mathcal{Q}$, $\mathcal{W}^\mathcal{K}$, and $\mathcal{W}^\mathcal{V}$ are the linear layers that transform the $\mathcal{Q}$, $\mathcal{K}$, and $\mathcal{V}$ respectively. The output with self-attention is then calculated as $\text{Softmax}( \frac{\mathcal{Q} \mathcal{K}^\intercal}{\sqrt{d}}) \mathcal{V}$, where $d$ denotes the embedding dimensions of $Q$ and $K$.
Efficient Parameter Transfer Learning (EPTL) aims to accelerate the training process on downstream datasets by transferring knowledge from pre-trained ViT models. Typically, only a small number of parameters are updated, while the rest are initialized from the pre-trained model and kept fixed during fine-tuning.

One EPTL example is the adapter approach \cite{houlsby2019parameter}, which involves training small additional modules (adapters) on top of a pre-trained base model. The adapter contains a few task-specific layers and has been successfully applied to various computer vision tasks, such as object detection and semantic segmentation \cite{chen2023visionadapter}. Given the input token $X$, the adapter $\mathcal{A}$ usually transforms the tokens as $X \leftarrow X + \mathcal{A}(X)$.

Another EPTL example is the Low-Rank Approximation (LoRA) method \cite{hu2021lora}, which approximates the weight increments of $W^\mathcal{Q}$ and $W^\mathcal{K}$ by $\Delta W^\mathcal{Q}$ and $\Delta W^\mathcal{K}$. During fine-tuning, $\Delta W^\mathcal{Q}$ and $\Delta W^\mathcal{K}$ are approximated by extra parameters, and updated via backward propagation, while $W^\mathcal{Q}$ and $W^\mathcal{K}$ are initialized from pre-trained models and fixed. Consequently, the query $\mathcal{Q} = XW^\mathcal{Q} + X\Delta W^\mathcal{Q}$ and $\mathcal{K} = XW^\mathcal{K} + X\Delta W^\mathcal{K}$.

Prompt tuning \cite{jia2022visual} is another EPTL example, in which input tokens of one or more layers are concatenated with learnable prompt tokens ${P}$. This combination of pre-trained models and few-shot learning enables rapid adaptation to new tasks. In the layer with prompts, $X \leftarrow [X, {P}]$. The tokens $X$ are learned from the fixed pre-trained model, while ${P}$ is trainable during fine-tuning.

In this work, we focus on developing a more advanced adapter by introducing spoofing-aware inductive bias to $\mathcal{A}$ when conducting token transformation. How to develop Prompt and LoRA for generalized FAS can be studied in the future. \textcolor{black}{A prior work that is related to our work is \cite{eccv2022vit}, which also adopts adapters for face anti-spoofing. However, \cite{eccv2022vit} only adopts simple multi-stream linear adapters for face anti-spoofing. The design of the adapters lacks insight from the data properties of the FAS task, such as locality, fine-grain information, style variances, etc. Our work incorporates the above insights in the design of our proposed S-Adapter.}

\section{Methodology}
In this section, we first provide preliminary knowledge about how to use adapters to fine-tune the vision transformer. Subsequently, we describe how our proposed S-Adapter is developed. Finally, we describe the final optimization method, which involves the proposed Token Style Regularization in the total loss function.
\subsection{ViT with S-Adapter}
\begin{figure}
    \centering
    \includegraphics[width=0.95\linewidth]{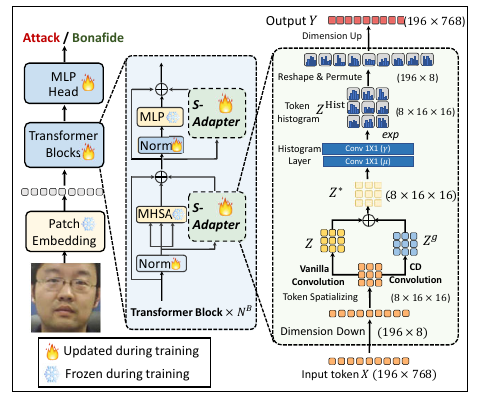}
    %\vspace{-2mm}
    \caption{The structure of ViT backbone and our S-Adapter. The S-Adapter is inserted into the ViT and updated during the training.  }
    \label{fig:framework}
    %\vspace{-2mm}
\end{figure}

Before delving into our S-Adapter, we first provide the necessary background on Vision Transformer (ViT) and adapters. In a ViT model \cite{ViT}, there can be a number of $N^B$ feed-forward transformer blocks, where the $i$-th block can be represented as $\mathcal{W}^{B}_i$. As depicted in Fig.~\ref{fig:framework}, each block usually contains a Multi-Head Self Attention (MHSA) layer $\mathcal{W}_i^{MSA}$ and a Multi-Layer Perceptron (MLP) layer $\mathcal{W}_i^{MLP}$, and each layer is accompanied by a Layer Normalization layer and a non-linear activation layer. By simplifying the skip connections, normalization layers, and activation layers, the inference procedure of each transformer block can be expressed as:
\begin{equation}\label{eq-ViT}
     {Y} = \mathcal{W}^{B}_i(X) = \mathcal{W}_i^{\text{MLP}} (\mathcal{W}^{\text{MSA}}_i(X)),
\end{equation}
where ${X}$ and ${Y}$ are the input and output tokens of the block respectively. 

In the traditional paradigm, ViT is pre-trained on a large-scale dataset, either by Self-Supervised Learning (SSL) or Supervised Learning (SL). Then, the pre-trained ViT is fine-tuned on a specific downstream task. Since fine-tuning the entire ViT model is challenging, recent studies on EPTL provide an efficient way of fine-tuning ViT by inserting into it small and task-specific adapter modules \cite{chen2023visionadapter, convpass}. With adapter modules, the inference process in the $i$-th block is turned from Eq.~\ref{eq-ViT} into 
\begin{equation}\label{eq-Adapter}
      {Y} =  \mathcal{A}_i^{\text{MLP}}(\mathcal{W}_i^{\text{MLP}}(\mathcal{A}_i^{\text{MSA}}(\mathcal{W}^{\text{MSA}}_i({X})),
\end{equation}
where $\mathcal{A}_i^{\text{MLP}}$ and $\mathcal{A}_i^{\text{MSA}}$ are the adapter modules after the MHSA and MLP layers respectively. During the fine-tuning, $\mathcal{W}_i^{MLP}$ and $\mathcal{W}_i^{MSA}$ are initialized by a pre-trained model, and they are fixed and not updated during the fine-tuning. On the other hand, the inserted adapter modules $\mathcal{A}^{\text{MSA}}$ and $\mathcal{A}^{\text{MLP}}$ are randomly initialized and updated by backward propagation. 

% Here needs more clarification.
Eq.~\ref{eq-Adapter} indicates the adapter's role in transforming token embeddings from the original space into a new space related to face anti-spoofing. However, vanilla adapters utilizing linear layers exhibit limitations in the embedding transformation for a face anti-spoofing (FAS) dataset. Firstly, linear-layer adapters lack image-specific inductive biases, such as spatial locality and 2D neighborhood structure \cite{convpass}, which are crucial for addressing FAS challenges due to the importance of visual local features \cite{CAI-2020-DRL}. \textcolor{black}{Beside, FAS classification needs more fine-grain information \cite{CDCN-CVPR-2020} and expects features to be robust against variations in imaging environments about illuminations and camera modules, which is non-trivial for linear layers to learn. }

We propose our S-Adapter to address the above limitations to adapt pre-trained ViT for generalized face anti-spoofing efficiently. As illustrated in Fig.~\ref{fig:inspiration}, our S-Adapter is inspired by traditional texture analysis methods in face anti-spoofing \cite{FAS-ColorTexture-TIFS-2016, deep-forest-lbp}. In the method of using local LBP features, the LBP descriptor is first used to extract the raw LBP maps, which are at a low level and sensitive to imaging conditions \cite{FAS-ColorTexture-TIFS-2016}. Then the LBP histogram is collected, improving the feature level with histogram statistical information. The feature representation with histogram can be more robust against lighting variations, which inspires us to introduce statistical information via the proposed S-Adapter.

The design of our S-Adapter is depicted in Fig.~\ref{fig:framework}, which links to the traditional LBP histogram method. The process of S-Adapter can be broken down into two main steps. The first step is \textbf{Token Map Extraction} and this step is analogous to the extraction of raw LBP maps. The second step is \textbf{Token Histogram Extraction}, in which we collect the histograms on token maps. This step is analogous to the LBP histogram feature extraction, improving the level of token embedding to be more robust against environmental changes. By incorporating the S-Adapter into the ViT architecture, we enhance its capability to handle cross-domain performance while maintaining efficiency by extracting statistical information.

\subsubsection{Token Map Extraction.} In the context of face anti-spoofing, local information is crucial for detection due to artifacts generated during the recapturing process, often appearing in local regions \cite{CAI-2020-DRL}. Since input tokens ${X}$ of a transformer block are flattened without the spatial structure. To extract local information, we reconstruct the 2D structure of $X$. Given the input token $X \in \mathbb{R}^{N^P \times C}$, where $N^P$ denotes the number of tokens and $C$ represents token embedding, we first reshape $X$ as $X^R \in \mathbb{R}^{H \times W \times C}$, with $H \times W = N^P$ (class token ignored). Then, we permute dimensions to obtain ${X}^{M} \in \mathbb{R}^{C\times H \times W}$. Consequently, tokens are represented in a 2D-image style, enabling the use of widely-used PyTorch-style 2D Convolution techniques for learning purposes.

With the spatial tokens ${X}^{M}$, we apply the 2D convolution $\mathcal{W}^{Conv}$ on it to extract the token map $Z$ in a learnable way that 
\begin{equation}
    Z = \mathcal{W}^{Conv}(X^{M}).
\end{equation}
Moreover, considering that the features of spoofing artifacts are often of fine-grain details, which can be represented by the gradients based on the Center Difference (CD), we extract token gradients based on the central difference \cite{CDCN-CVPR-2020} of tokens. The gradient of a token $Z_n$ can be represented as
\begin{equation}
      Z^{g}_n  = \sum_{p\in \mathcal{P}^n}\omega(p)\cdot (Z_{p} - Z_n),
\end{equation}
where $Z_n$ is the $n$-th element of $Z$ ($n<N^P$), $\mathcal{P}^n$ is the index sets of the spatial neighbors of tokens of $Z_n$, $\omega$ is the kernel weight of $\mathcal{W}^{Conv}$. With the token gradients $Z^{g}$, the final token maps $Z^*$ can be calculated as 
\begin{equation}
Z^*= (1-\theta) Z  + \theta Z^{g},
\end{equation}
where $\theta$ is a constant ratio that balances the localized transformed tokens and the gradient of tokens. %We use $\theta=0.7$ as suggested in \cite{CDCN-CVPR-2020}.

\subsubsection{Token Histogram Extraction.} \textcolor{black}{The token map $Z^*$ contains fine-grain texture features but remains sensitive to domain shifts. To address this issue, we propose to gather token statistics through the computation of token histograms $Z^{Hist}$ to mitigate the domain shift problem, as statistical information can improve the representation capability \cite{FAS-ColorTexture-TIFS-2016} for the FAS task. The histograms are computed for each token of the token map $Z^*$. In each transformer layer, there are $14\times14$ patch tokens. Therefore, $14\times14$ histograms are calculated, and one histogram for one patch token respectively. }

The token histogram calculation process involves: 1) segmenting the feature range into discrete bins, and 2) enumerating the occurrences of each value within their respective bins. However, these operations are non-differentiable, which obstructs gradient calculation and prevents the model from being updated during backward propagation. To fit into the modern fashion of deep learning, we utilize the soft binning \cite{HistTexture} and extract differentiable token histograms.

By defining the bin center and the bin width as $\mu$ and $\gamma$, respectively, given the token map $Z^* \in \mathbb{R}^{C \times H \times W} $, the soft binned histogram $Z^{Hist}$ is defined as  
\begin{equation}\label{eq-soft-bin}
    Z^{Hist}_{chw} = \frac{1}{JK} \sum_{j=1}^J \sum_{k=1}^K e^{-\gamma_c^2 (Z^*_{c, h+j, w+k} - \mu_c)^2}, 
    \end{equation}
where $c$ denotes the channel dimension, $h$ and $w$ are spatial dimensions of the token histogram, $J$ and $K$ are the spatial sizes of the sliding window. To keep the size of tokens unchanged, $J=K=3$, the stride of the window is 1, and the padding size is 1. Eq.~\ref{eq-soft-bin} is differentiable and the bin center ($\mu$) and bin width ($\gamma$) are also trainable and can be updated during the backward propagation. To learn $\mu$ and $\gamma$, Eq.~\ref{eq-soft-bin} can be dismantled as 
\begin{equation}\label{eq-dis}
\begin{aligned}
      Z^{Hist}_{chw} &= \frac{1}{JK} \sum_{j=1}^J \sum_{k=1}^K e^{-U_c^2},\\
     U_c &= \gamma_c (Z^*_{c, h+j, w+k} - \mu_c),
\end{aligned}
\end{equation}

where $U_c$ denotes the $c$-th element of a vector $U$. $U$ can be learned from two consecutive pixel-wise convolutional layers. In detail, we define $\mathcal{W}^{Conv1}$, a $C$-channel pixel-wise $1\times 1$ convolution of which the kernel weight is fixed as 1 and the bias is learnable, which can serve as $\mu$. In this way, $\mathcal{W}^{Conv1}(Z)$ leads to $Z-\mu$.
Likewise, we can define $\mathcal{W}^{Conv2}$, a $C$-channel pixel-wise convolution of which the bias 0, and the learnable kernel weight can serve as $\gamma$. In this way,  $\mathcal{W}^{Conv2}(X)$ leads to $\gamma X$. As a result,
\begin{equation}
\begin{aligned}
    U & = \mathcal{W}^{Conv2}(\mathcal{W}^{Conv1}(Z^*_{:, h+j, w+k})),\\
    U_c & = \gamma_c (Z^*_{c, h+j, w+k}-\mu_c).
\end{aligned}
\end{equation}

\begin{figure}
    \centering
    \includegraphics[width=\linewidth]{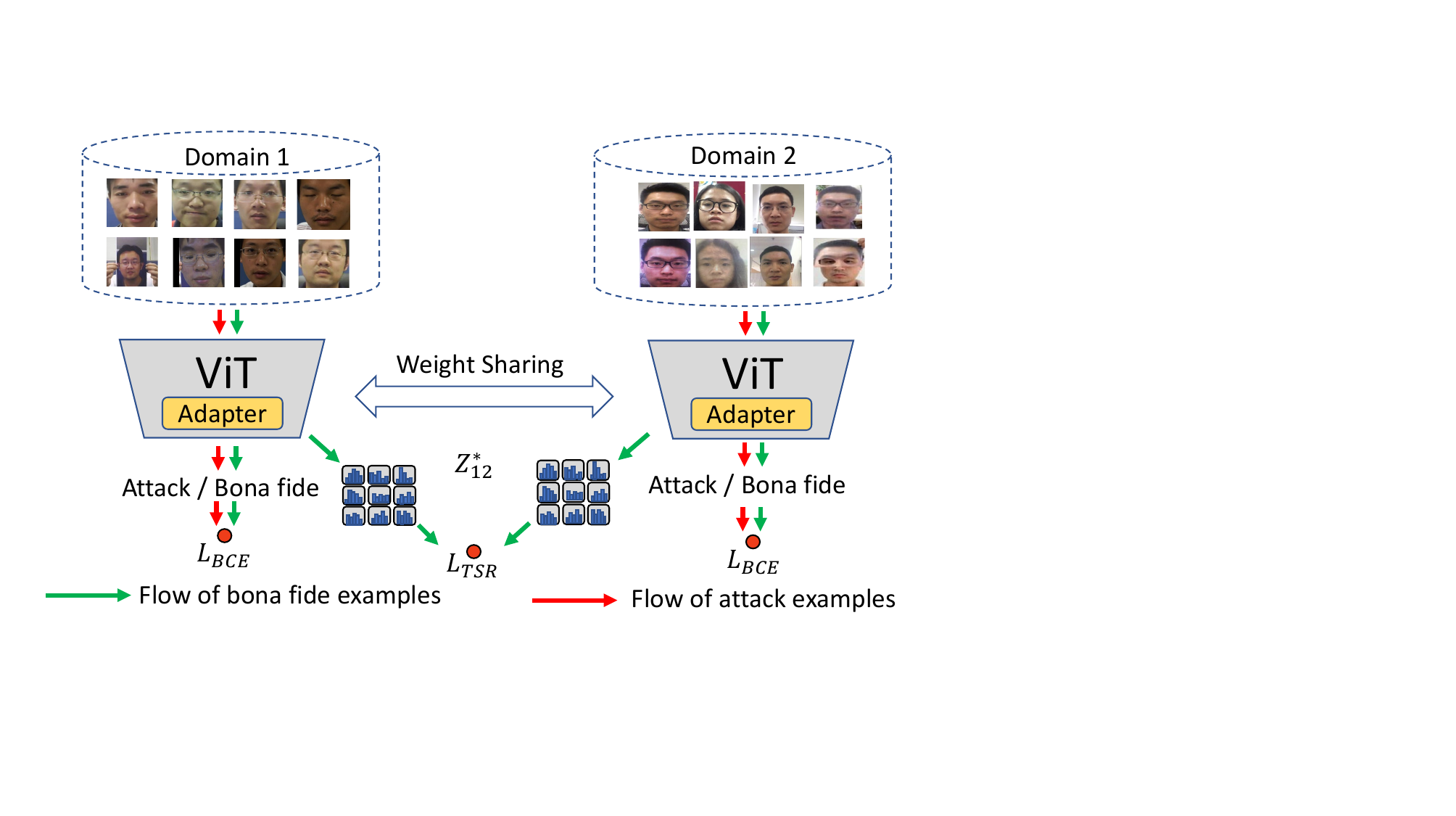}  
      %\vspace{-5mm}
    \caption{The overall optimization. The bona fide and attack examples are used to calculate the binary cross-entropy loss ($\mathcal{L}_{BCE}$), but only the bona fide examples are used to calculate the Token Style Regularization ($\mathcal{L}_{TSR}$).}\label{fig:loss}
   %\vspace{-4mm}
\end{figure}

\textcolor{black}{The output token histogram $Z^{Hist} \in \mathcal{R}^{C\times H \times W}$ is finally reshaped and permuted as the output token (${N^P \times 8}$). Then, it is projected to $Y \in \mathbb{R}^{N^P \times 768}$ via a linear layer, which can achieve dimensional and semantic alignment for the fusion with the original tokens.  }

\subsection{Token Style Regularization}
\textcolor{black}{The S-Adapter extracts statistical tokens, but the generalization of the token histogram is still affected by the variances of color and lighting, which can be summarized as style variances. To alleviate the adverse effect of learning domain-variant style components when extracting the tokens and enhance cross-domain generalization performance.} Existing deep learning techniques for face anti-spoofing involve learning features from data, which can be categorized into content and style features \cite{SSAN}. Content features pertain to artifacts or cues crucial to the classification and are expected to be domain-invariant. By contrast, the style features encompass low-level image attributes, which are sensitive to variations in environmental changes. Diverse capturing conditions yield distinct styles, which contribute to domain shifts. Consequently, minimizing the style features learned from data is desired for domain-invariant features. Prior studies have explored disentanglement learning to separate content and style features through adversarial learning or other intricate mechanisms \cite{liu2020disentangling}. However, adversarial learning is often unstable, prompting us to propose TSR, a straightforward approach that does not require complex adversarial training.

Since that image style can be represented by the Gram matrix \cite{eccv2016-perceptualloss} of images, we formulate TSR based on the Gram matrix to minimize domain-variant styles. 
Given a token map $Z \in \mathbb{R}^{C \times H \times W}$, the entry ($k$, $k'$) of its gram matrix can be defined as  
    \begin{equation}
       G(Z)_{k, k'} = \frac{1}{C H W} \sum_{h=1}^H \sum_{w=1}^W Z_{k, h,w} Z_{k',h,w},
    \end{equation}
where $Z_{k,h,w}$ denotes the element of $Z$ at the coordiates of ($h, w$) at $k$-th channel.
Considering that all bona fide examples can be regarded from one domain \cite{SSDG-CVPR-2020}, and thus their style features should be regularized to be the same. Thus, we propose our single-side Token Style Regularization, which minimizes the style variance between bona fide examples from different domains, as depicted in Fig.~\ref{fig:loss}. Given tokens of bona fide examples from two domains (\textit{e.g.}, ${Z}^{D1}$, ${Z}^{D2}$), the TSR $\mathcal{L}_{TSR}$ is represented as 
 \begin{equation}
     \mathcal{L}_{TSR} = ||G({Z}^{D1}) - G({Z}^{D2})||^2_F.
 \end{equation}

\subsection{Final Optimization}
In the final optimization, we utilize the binary cross-entropy loss $L_{BCE}$, as the face anti-spoofing can be regarded as a binary classification problem, where the label ``0'' denotes the bonafide example and label ``1'' denotes spoofing attack example. When the number of domains is more than 2, we enumerate every two domains to calculate $\mathcal{L}_{TSR}$ and then calculate the average of the $\mathcal{L}_{TSR}$ to obtain $\mathcal{L}^a_{TSR}$. As such, the final loss can be represented as 
\begin{equation}
    \mathcal{L}_{total} = \mathcal{L}_{BCE} + \lambda \mathcal{L}^a_{TSR},
\end{equation}
where $\lambda$ is a constant scaling factor.  When calculating $\mathcal{L}^a_{TSR}$, we use the $Z$ from the last transformer block to calculate the gram matrix.

\section{Experiment}
\subsection{Datasets, Protocols, and Implementations}
Our experiments involve the use of several benchmark datasets, including CASIA-FASD \cite{DB-CASIAFASD}, IDIAP REPLAY ATTACK \cite{LBP-FAS-BIOSIG-2012}, MSU MFSD \cite{FAS-IDA-TIFS-2015}, OULU-NPU \cite{OULU_NPU_2017}, and SiW-M \cite{SiW-M-2019}. To evaluate our models, we employ Half-Total Error Rate (HTER), Attack Classification Error Rate (ACER), Equal Error Rate (EER), Area Under Receiver Operating Characteristic Curve (AUC), and the True Positive Rate (TPR) when False Positive Rate (FPR) equals $1\%$ ($TPR@FPR=1\%$). These rigorous evaluation procedures ensure the reliability and validity of our findings.

To conduct our experiments, we utilized the PyTorch 1.9 framework and performed training and testing on a single NVIDIA GTX 2080 Ti GPU. We follow \cite{george2021effectiveness, eccv2022vit} to use ViT-Base as the ViT backbone. In data processing, we utilized MTCNN \cite{MTCNN} to detect faces and resized the cropped face images to $224\times224$ as the input to the ViT-Base. During training, we employed the Adam optimizer with an initial learning rate of 0.0001.  %To enhance the robustness of our models, we applied various data augmentation techniques, including random flipping, random cropping, random patch shuffle, and random color jitter. These methods helped to mitigate overfitting and improve the generalization ability of our models. 

\begin{table}[tbp]
  \centering
  \caption{\textcolor{black}{Intra-domain experiments on the four protocols (P1,P2,P3,P4) of the OUPU-NPU datasets. } }
    \begin{tabular}{|c|c|c|c|c|}
    \hline
    Protocol & Method & \multicolumn{1}{l|}{APCER(\%)} & \multicolumn{1}{l|}{BPCER(\%)} & \multicolumn{1}{l|}{ACER(\%)} \bigstrut\\
    \hline
    \multirow{6}[12]{*}{P1} & Auxiliary\cite{FAS-Auxiliary-CVPR-2018} & 1.6   & 1.6   & 1.6 \bigstrut\\
\cline{2-5}          & DCL-FAS \cite{CAI-2020-DRL} & 5.4   & 4.0     & 4.7 \bigstrut\\
\cline{2-5}          & Spoof-Trace \cite{liu2020disentangling} & 0.8   & 1.3     & 1.1 \bigstrut\\
\cline{2-5}          & CDCN \cite{CDCN-CVPR-2020}  & 0.4   & 1.7   & 1 \bigstrut\\
\cline{2-5}          & CDCN++ \cite{CDCN-CVPR-2020} & 0.4   & \textbf{0}     & 0.2 \bigstrut\\
\cline{2-5}          & EPCR \cite{EPCR2023}  & 1.7   & \textbf{0}     & 0.8 \bigstrut\\
\cline{2-5}          & S-Adapter-TSR (Ours)  & \textbf{0.05}  & 0.15  & \textbf{0.1} \bigstrut\\
    \hline
    \multirow{6}[12]{*}{P2} & Auxiliary\cite{FAS-Auxiliary-CVPR-2018} & 2.7   & 2.7   & 2.7 \bigstrut\\
\cline{2-5}          & DCL-FAS \cite{CAI-2020-DRL} & 3.4   & \textbf{0.1}   & 1.9 \bigstrut\\
\cline{2-5}          & Disentangle\cite{liu2020disentangling} & 2.3   & 1.6     & 1.9 \bigstrut\\
\cline{2-5}          & Spoof-Trace \cite{liu2020disentangling} & 2.3   & 1.6     & 1.9 \bigstrut\\
\cline{2-5}          & CDCN \cite{CDCN-CVPR-2020}  & 1.5   & 1.4   & 1.5 \bigstrut\\
\cline{2-5}          & CDCN++ \cite{CDCN-CVPR-2020} & 1.8   & 0.8   & 1.3 \bigstrut\\
\cline{2-5}          & EPCR \cite{EPCR2023}  & 0.8   & 0.4   & \textbf{0.6} \bigstrut\\
\cline{2-5}          & S-Adapter-TSR (Ours) & 1.2   & 0.4   & 0.8 \bigstrut\\
    \hline
    \multirow{6}[12]{*}{P3} & Auxiliary\cite{FAS-Auxiliary-CVPR-2018} & 2.7±1.3 & 3.1±1.7 & 2.9±1.5 \bigstrut\\
\cline{2-5}          & DCL-FAS \cite{CAI-2020-DRL}& 4.6±3.6 & \textbf{1.3}±1.8 & 3.0±1.5 \bigstrut\\
\cline{2-5}          & Spoof-Trace \cite{liu2020disentangling} & 1.6±1.6   & 4.0±5.4     & 2.8±3.3 \bigstrut\\
\cline{2-5}          & CDCN \cite{CDCN-CVPR-2020} & 2.4±1.3 & 2.2±2.0 & 2.3±1.4 \bigstrut\\
\cline{2-5}          & CDCN++\cite{CDCN-CVPR-2020} & 1.7±1.5 & 2.0±1.2 & 1.8±0.7 \bigstrut\\
\cline{2-5}          & EPCR \cite{EPCR2023}  & \textbf{0.4}±0.5 & 2.5±3.8 & 1.5±2.0 \bigstrut\\
\cline{2-5}          &  S-Adapter-TSR (Ours)  & 0.4.±0.6 & 1.8±1.2 & \textbf{1.1}±1.0 \bigstrut\\
    \hline
    \multirow{6}[12]{*}{P4} & Auxiliary\cite{FAS-Auxiliary-CVPR-2018} & 9.3±5.6 & 10.4±6.0 & 9.5±6.0 \bigstrut\\
\cline{2-5}          & DCL-FAS \cite{CAI-2020-DRL} & 8.1±2.7 & 6.9±5.8 & 7.2±3.9 \bigstrut\\
\cline{2-5}          & Spoof-Trace \cite{liu2020disentangling} & 2.3+3.6   & 5.2±5.4     & 3.8±4.2 \bigstrut\\
\cline{2-5}          & CDCN \cite{CDCN-CVPR-2020} & 4.6±4.6 & 9.2±8.0 & 6.9±2.9 \bigstrut\\
\cline{2-5}          & CDCN++ \cite{CDCN-CVPR-2020} & 4.2±3.4 & 5.8±4.9 & 5.0±2.9 \bigstrut\\
\cline{2-5}          & EPCR \cite{EPCR2023}  & \textbf{0.8}±2.0 & 7.5±11.7 & 3.3±4.9 \bigstrut\\
\cline{2-5}          &  S-Adapter-TSR (Ours)  & 1.5±3.1 & \textbf{3.9}±4.6 & \textbf{2.7}±3.5 \bigstrut\\
    \hline
    \end{tabular}%
  \label{tab:oulu}%
\end{table}

\begin{table*}[tbp]
  \centering
  \caption{Experimental results on the leave-one-out benchmark MICO. Results are in terms of HTER (\%) and AUC (\%).}
  %\vspace{-0.8em}
    \resizebox{0.95\textwidth}{!}{
    \begin{tabular}{l|c|c|c|c|c|c|c|c|c}
    \hline
    \hline
    \multicolumn{1}{c|}{\multirow{2}[4]{*}{Method}}& \multirow{2}[4]{*}{Venue} & \multicolumn{2}{c|}{C\&I\&O to M} & \multicolumn{2}{c|}{O\&M\&I to C} & \multicolumn{2}{c|}{O\&C\&M to I} & \multicolumn{2}{c}{I\&C\&M to O} \bigstrut\\
\cline{3-10}   &   & HTER(\%) & AUC(\%) & HTER(\%) & AUC(\%) & HTER(\%) & AUC(\%) & HTER(\%) & AUC(\%) \\
    \hline
    MMD-AAE \cite{MMDAAE-CVPR-2018} & CVPR 2018 &  27.08 & 83.19 & 44.59 & 58.29 & 31.58 & 75.18 & 40.98 & 63.08\\
    \hline 
    MADDG \cite{MADDG-CVPR-2019} & CVPR 2019  & 17.69 & 88.06 & 24.50 & 84.51 & 22.19 & 84.99 & 27.98 & 80.02\\
    \hline
    RFMetaFAS \cite{RFMetaFAS-AAAI-2020}&AAAI 2020   & 13.89      &  93.98     &  20.27     &     88.16  &    17.30   &  90.48     &  16.45     &   91.16\\
    \hline
    NAS-Baesline \cite{NASFAS-TPAMI-2020}& T-PAMI 2021 & 11.62 & 95.85 & 16.96 & 89.73 & 16.82 & 91.68 & 18.64 & 88.45 \\
    \hline
    NAS w/ D-Meta \cite{NASFAS-TPAMI-2020} & T-PAMI 2021 & 16.85 & 90.42 & 15.21 & 92.64 & 11.63 & 96.98 & 13.16 & 94.18 \\
    \hline
    NAS-FAS \cite{NASFAS-TPAMI-2020}& T-PAMI 2021 & 19.53 & 88.63 & 16.54 & 90.18 & 14.51 & 93.84 & 13.80 & 93.43 \\
    \hline
    SSDG-M  \cite{SSDG-CVPR-2020}& CVPR 2020 &16.67 & 90.47 & 23.11 & 85.45 & 18.21 & 94.61 & 25.17 & 81.83 \\
     \hline
    SSDG-R  \cite{SSDG-CVPR-2020} & CVPR 2020 & 7.38  & 97.17 & 10.44 & 95.94 & 11.71 & 96.59 & 15.61 & 91.54 \\
    \hline
    FAS-DR-BC(MT)  \cite{MetaTeacher-TPAMI-2021} &T-PAMI 2021   &     11.67    &  93.09     &  18.44   &  89.67     &     11.93  &    94.95   &  16.23     & 91.18 \\
    \hline
    SSAN-R  \cite{SSAN} & CVPR 2022 & 6.57 & 98.78 & 10.00 & 96.67 &  8.88 & 96.79 & 13.72 & 93.62 \\
    \hline
    PatchNet  \cite{wang2022patchnet}&CVPR 2022  & 7.10 & 98.46 & 11.33 & 94.58 &  13.4 & 95.67 & 11.82 & 95.07 \\ \hline
    AMEL\cite{zhou2022adaptive}& ACM MM 2022 & 10.23 & 96.62 & 11.88 & 94.39 &  18.60 & 88.79 & 11.31 & 93.36\\  \hline
    MetaPattern \cite{metapattern} & T-IFS 2022  & 5.24  &	97.28 &	9.11 &	96.09 &	15.35 &	90.67 & 12.40 &	94.26 \\
    \hline
     ViT$\dagger$ \cite{eccv2022vit} &ECCV 2022 & 4.75 & 98.79 & 15.70 & 92.76 & 17.68 & 86.66 & 16.46 & 90.37 \\
    \hline
    %ViT-Convpass  & 8.80 & 96.89 & 8.50 & 96.33 & 20.71 & 86.75 & 13.36 & 93.53 \\
   %ViT-Base-S-Adapter (Ours)  &  &  &  &  &  &  &  &  \\
    \hline
    ViT-S-Adapter-TSR  & Ours& \textbf{3.43}  & \textbf{99.50} & \textbf{6.32} & \textbf{97.82} & \textbf{7.16} & \textbf{97.61} & \textbf{7.21} & \textbf{98.00} \\
    
    \hline
    \hline

    \end{tabular}%
    }
  \label{tab:sota-mico}%
\end{table*}%

\begin{table}[tbp]
  \centering
  %\vspace{-3mm}
  \caption{Experimental results with limited source domains. Results are in terms of HTER (\%) and AUC (\%). }   \label{tab:sota-MI-CO}%
  %\vspace{-0.8em}
  \resizebox{0.45\textwidth}{!}{
    \begin{tabular}{l|c|c|c|c|c}
    \hline
    \hline
    \multirow{2}[4]{*}{Method} & \multirow{2}[4]{*}{Venue} & \multicolumn{2}{c|}{M\&I to C} & \multicolumn{2}{c}{M\&I to O} \\
\cline{3-6}    &  & HTER(\%) & AUC(\%) & HTER(\%) & AUC(\%) \\
    \hline
MS-LBP \cite{MicroTexture-IJCB-2011} & IJCB 2011 & 51.16 & 52.09 & 43.63 & 58.07 \\
\hline
IDA \cite{FAS-IDA-TIFS-2015} & T-IFS 2015& 45.16 & 58.80 & 54.52 & 42.17\\
\hline
ColorTexture \cite{FAS-ColorTexture-TIFS-2016} & T-IFS 2016 & 55.17 & 46.89 &  53.31 & 45.16 \\\hline
LBP-TOP \cite{LBP-TOP-EJIVP-2014} &EJIVP 2014 & 45.27 &  54.88 & 47.26 & 50.21 \\\hline
MADDG \cite{MADDG-CVPR-2019}& CVPR 2019 & 41.02 & 64.33 & 39.35 & 65.10  \\
\hline
SSDG-M \cite{SSDG-CVPR-2020} & CVPR 2020 & 31.89 & 71.29 & 36.01 & 66.88  \\
\hline
AMEL\cite{zhou2022adaptive}&ACM MM 2022 & 23.33& 85.17 &	\textbf{19.68} & \textbf{87.01} \\
   \hline
SSAN-M \cite{SSAN}&CVPR 2022& 30.00 &76.20 &29.44 & 76.62 \\
   \hline
MetaPattern\cite{metapattern} &T-IFS 2022& 30.89 & 72.48 &	20.94
& 86.71 \\
    \hline
%ViT-HistA (Ours)& 19.37  &  85.70& 22.60 &  82.75 \\
ViT-S-Adapter-TSR & Ours & \textbf{17.93}&	\textbf{89.56}&	\underline{19.76}&	\underline{86.87}\\
\hline
    \hline
    \end{tabular}%
    }
  %\vspace{-4mm}
\end{table}%

%\vspace{-2mm}
\subsection{Intra-Domain Evaluation}
\textcolor{black}{We first report the intra-domain experiment by using the OULU-NPU dataset's four protocols \cite{OULU_NPU_2017}, and the experimental results are in Table~\ref{tab:oulu}, and the used metrics are Attack Presentation Classification Error Rate (APCER), Bona Fide Presentation Classification Error Rate (BPCER), and Average Classification Error Rate (ACER). The ACER is the average of APCER and BPCER. Compared with the state-of-the-art methods, our method shows prominent performance, as it generally achieves the best ACER compared with other methods.  Thus, our method is effective in the intra-dataset experiment.}
\subsection{Cross-Domain Evaluation}
\subsubsection{Leave-one-out cross-domain benchmark} 
To begin with, we compare our proposed method with state-of-the-art methods on the leave-one-out cross-domain benchmark \cite{MADDG-CVPR-2019}, which consists of the CASIA-FASD (\textit{C}) \cite{DB-CASIAFASD}, IDIAP REPLAY ATTACK (\textit{I}) \cite{LBP-FAS-BIOSIG-2012}, MSU MFSD (\textit{M}) \cite{FAS-IDA-TIFS-2015}, and OULU-NPU (\textit{O}) \cite{OULU_NPU_2017} datasets. This benchmark can also be referred to as the MICO benchmark \cite{metapattern} and has been widely used for cross-domain performance evaluation \cite{SSDG-CVPR-2020, SSAN, eccv2022vit, NASFAS-TPAMI-2020, MetaTeacher-TPAMI-2021, RFMetaFAS-AAAI-2020}. We follow the MICO benchmark's protocols described in \cite{MADDG-CVPR-2019} and present our HTER and AUC results in Table~\ref{tab:sota-mico}.

\begin{table*}[htbp]
  \centering
  \caption{The results of the 5-shot cross-domain experiment. 5 bona fide examples and 5 attack examples from the target domain are used to fine-tune the pre-trained model. The results are in terms of HTER (\%), AUC(\%) TPR(\%)@FPR=1\%}
  \resizebox{.9\textwidth}{!}{
    \begin{tabular}{l|ccc|ccc|ccc|ccc}
    \hline
    \hline
    \multicolumn{1}{c}{\multirow{2}[4]{*}{Method}} & \multicolumn{3}{c|}{C\&I\&O to M} & \multicolumn{3}{c|}{O\&M\&I to C} & \multicolumn{3}{c|}{C\&M\&O to I} & \multicolumn{3}{c}{C\&I\&M to O} \bigstrut\\
\cline{2-13}    \multicolumn{1}{c}{} & \multicolumn{1}{c|}{HTER} & \multicolumn{1}{c|}{AUC} & TPR@FPR=1\% & \multicolumn{1}{c|}{HTER} & \multicolumn{1}{c|}{AUC} & TPR@FPR=1\% & \multicolumn{1}{c|}{HTER} & \multicolumn{1}{c|}{AUC} & TPR@FPR=1\% & \multicolumn{1}{c|}{HTER} & \multicolumn{1}{c|}{AUC} & TPR@FPR=1\% \bigstrut\\
    \hline
    %ViT$\dagger$ (ECCV-2022)\cite{eccv2022vit} &       &       &       &       &       &       &       &       &       &       &       &  \bigstrut[t]\\
    ViTAF$\dagger$ (ECCV-2022) \cite{eccv2022vit} & 4.75  & 98.59 &   80.00    & 4.19  & 98.59 &  57.86     & 3.28  & 99.27 &  76.92     & 10.74 & 95.70  & 51.13 \\
    \hline
    ViT-S-Adapter (Ours) & 2.38 & 99.72 &   94.67    & 3.82 & 99.52&  78.12   & 1.76 & 99.81 &  94.75   &    4.24    &99.21 & 85.12 \bigstrut[b]\\
    \hline
    \hline
    \end{tabular}%
    }
  \label{tab:5-shot}%
\end{table*}%

\begin{table*}[t]
\caption{Results of LOO protocols on SiW-M dataset \cite{SiW-M-2019}. The values ACER(\%) reported on testing sets are obtained with the threshold of 0.5. The best results are bolded.}
%\vspace{-1.2em}
  \centering
   \resizebox{18cm}{!}{
    \begin{tabular}{cccccccccccccccc}
    \hline
    \hline
\textbf{Method} & \textbf{Metric(\%)} & \textbf{Replay} & {\textbf{Print}} & \multicolumn{5}{c}{\textbf{Mask Attacks}} & \multicolumn{3}{c}{\textbf{Makeup Attacks}} & \multicolumn{3}{c}{\textbf{Partial Attacks}} & \textbf{Average} \\
    \cline{5-15} &  &       &       & \textbf{Half} & \textbf{Silicone} & \textbf{Trans} & \textbf{Paper} & \textbf{Manne} & \textbf{Obfusc} & \textbf{Imperson} & \textbf{Cosmetic} & \textbf{Funny Eye} & \textbf{Paper Glasses} & \textbf{Partial Paper}& \\
    \midrule[2pt]
    %\multirow{2}{*}{{SVM+LBP}\cite{OULU_NPU_2017}} & ACER  & 20.6 & 18.4  & 31.3  & 21.4  & 45.5 & 11.6  & 13.8  & 59.3 & 23.9  & 16.7 & 35.9 & 39.2 & 11.7  & 26.9\pm14.5 \\
    %&EER  & 20.8 & 18.6  & 36.3  & 21.4  & 37.2 & 7.5  & 14.1  & 51.2 & 19.8  & 16.1  & 34.4 & 33.0 & 7.9  & 24.5\pm12.9\\
    %\hline
    \multirow{2}{*}{{Auxiliary}\cite{FAS-Auxiliary-CVPR-2018}} & ACER  & 16.8 & 6.9 & 19.3 & 14.9 & 52.1 & 8.0 & 12.8 & 55.8 & 13.7 & 11.7 & 49.0 & 40.5 & 5.3  & 23.6$\pm$18.5 \\
    &EER  & 14.0 & 4.3 & 11.6 & 12.4 & 24.6 & 7.8 & 10.0 & 72.3 & 10.1 & 9.4 & 21.4 & 18.6 & 4.0   & 17.0$\pm$17.7\\
    \hline
    \multirow{2}{*}{{BCN}\cite{yu2020material}} & ACER & 12.8 & 5.7 & 10.7 & 10.3 & 14.9 & 1.9 & 2.4 & 32.3 & 0.8 & 12.9 & 22.9 & 16.5 & 1.7 & 11.2$\pm$9.2  \\
    &EER  & 13.4 & 5.2 & 8.3 & 9.7 & 13.6 & 5.8 & 2.5 & 33.8 & \textbf{0.0} & 14.0 & 23.3 & 16.6 & 1.2 & 11.3$\pm$9.5 \\
    \hline
    \multirow{2}{*}{{CDCN++}\cite{CDCN-CVPR-2020}} & ACER  & 10.8 & 7.3 & 9.1 & 10.3 & 18.8 & 3.5 & 5.6 & 42.1 & 0.8 & 14.0 & 24.0 & 17.6 & 1.9 & 12.7$\pm$11.2  \\
    &EER  & 9.2 & 5.6 & 4.2 & 11.1 & 19.3 & 5.9 & 5.0 & 43.5 & \textbf{0.0} & 14.0 & 23.3 & 14.3 & \textbf{0.0} & 11.9$\pm$11.8\\
    \hline
    \multirow{2}{*}{{DC-CDN}\cite{DCCDN-IJCAI-2021}} & ACER & 12.1 & 9.7 & 14.1 & 7.2 & 14.8 & 4.5 & 1.6 & 40.1 & 0.4 & 11.4 & 20.1 & 16.1 & 2.9  &11.9$\pm$10.3 \\
    &EER & 10.3 & 8.7 & 11.1 & 7.4 & 12.5 & 5.9 & 0.0 & 39.1 & \textbf{0.0} & 12.0 & 18.9 & 13.5 & 1.2 &  10.8$\pm$10.1 \\
    \hline
    %\multirow{2}{*}{{SpoofTrace}\cite{liu2020disentangling}} & ACER & 7.6 & \textbf{3.8} & 8.4 & 13.8 & 14.5 & 5.3 & 4.4 & 35.4 & \textbf{0.0} & 19.3 & 21.0 & 20.8 & 1.6 &  14.2$\pm$13.2   \\
    %&EER  & 13.4 & 5.2 & 8.3 & 9.7 & 13.6 & 5.8 & 2.5 & 33.8 & \textbf{0.0} & 14.0 & 23.3 & 16.6 & 1.2 & 12.0$\pm$10.0 \\
    %\hline
     \multirow{2}{*}{{SpoofTrace}\cite{liu2020disentangling}} & ACER & 7.8 & 7.3 & 7.1 & 12.9 & 13.9 & 4.3 & 6.7 & 53.2 & 4.6 & 19.5 & 20.7 & 21.0 & 5.6 &  14.2$\pm$13.2   \\
    &EER  & \textbf{7.6} & 3.8 & 8.4 & 13.8 & 14.5 & 5.3 & 4.4 & 35.4 & \textbf{0.0} & 19.3 & 21.0 & 20.8 & 1.6 & 12.0 $\pm$ 10.0 \\
         \hline
    \multirow{2}{*}{{DTN}\cite{DTL}} & ACER & 9.8 & 6.0 & 15.0 & 18.7 & 36.0 & 4.5 & 7.7 & 48.1 & 11.4 & 14.2 & 19.3 & 19.8 & 8.5 & 16.8$\pm$11.1    \\
    &EER & 10.0 & \textbf{2.1} & 14.4 & 18.6 & 26.5 & 5.7 & 9.6 & 50.2 & 10.1 & 13.2 & 19.8 & 20.5 & 8.8 & 16.1$\pm$12.2  \\
    \hline
    \multirow{2}{*}{{DTN(MT)}\cite{MetaTeacher-TPAMI-2021}} & ACER & 9.5 & 7.6 & 13.1 & 16.7 & 20.6 & 2.9 & 5.6 & 34.2 & 3.8 & 12.4 & 19.0 & 20.8 & 3.9 & 13.1$\pm$8.7   \\
    &EER & 9.1 & 7.8 & 14.5 & 14.1 & 18.7 & 3.6 & 6.9 & 35.2 & 3.2 & 11.3 & 18.1 & 17.9 & 3.5 &  12.6$\pm$8.5 \\
    \hline
    \multirow{2}{*}{{FAS-DR(Depth)}\cite{MetaTeacher-TPAMI-2021}} & ACER & 7.8 & 5.9 & 13.4 & 11.7 & 17.4 & 5.4 & 7.4 & 39.0 & 2.3 & 12.6 & 19.6 & 18.4 & 2.4 &  12.6$\pm$9.5  \\
    &EER & 8.0 & 4.9 & 10.8 & 10.2 & 14.3 & 3.9 & 8.6 & 45.8 & 1.0 & 13.3 & 16.1 & 15.6 & 1.2 &  11.8$\pm$11.0 \\
    \hline
    \multirow{2}{*}{{FAS-DR(MT)}\cite{MetaTeacher-TPAMI-2021}} & ACER & \textbf{6.3} & 4.9 & 9.3 & 7.3 & 12.0 & 3.3 & 3.3 & 39.5 & 0.2 & 10.4 & 21.0 & 18.4 & 1.1  &10.5$\pm$10.3 \\
    &EER & 7.8 & 4.4 & 11.2 & 5.8 & 11.2 & 2.8 & 2.7 & 38.9 & 0.2 & 10.1 & 20.5 & 18.9 & 1.3 &  10.4$\pm$10.2 \\
    \hline
    \multirow{2}{*}{ViT\cite{liao2023domain}} & ACER  & 11.35 & 5.58  & 3.44  & 9.63  & 16.73 & 1.47  & 2.89  & 26.60 & 1.90  & 9.04  & 23.14 & 11.23 & 2.44  & 9.65$\pm$8.19 \\
    &EER  & 11.18 & 7.32  & 3.89  & 9.63  & 14.32 & 0.00  & 3.50  & 23.48 & 1.64  & 9.20  & 20.38 & 11.32 & 1.86  & 9.06$\pm$7.21\\
    %\hline
   % \multirow{2}{*}{ViTranZFAS (FC)  \cite{liao2023domain}} & ACER  &  &   &  &   &  &   &  & &   &   & &  &   & \\
    %&EER  &  &  &    &  &   &   & &   &   &  &  &   & \\
     %\midrule[2pt]
        \hline
    \multirow{2}{*}{ViT-S-Adapter (Ours)} & ACER  & 8.93 & \textbf{4.08}  & \textbf{1.81} & \textbf{2.02}  & \textbf{1.61} & \textbf{0.39}  &  \textbf{0.62} & \textbf{4.00} & 1.09 & \textbf{6.60} & \textbf{13.09} & \textbf{0.54}  & \textbf{0.43}  & \textbf{3.48}$\pm$3.90 \\
    & EER & \textbf{5.38} & 3.48 &  \textbf{1.67} & \textbf{2.96} & \textbf{1.36} & \textbf{0.00}  & \textbf{0.00}  & \textbf{4.35} & \textbf{0.00} & \textbf{7.20} & \textbf{10.25} & \textbf{0.48}  & 0.23  & \textbf{2.87}$\pm$3.20  \\
    \hline
    \hline
    \end{tabular}
    }
  \label{tab:intra_SiWM}%
  %\vspace{-2mm}
\end{table*}%  
We conducted a fair comparison by extracting the results of ViT$\dagger$ from \cite{eccv2022vit} without using any supplementary data from the CelebA-Spoof dataset \cite{CelebA-Spoof}. ViT$\dagger$ utilizes the same ViT-Base backbone as our proposed approach. As presented in Table~\ref{tab:sota-mico}, our method outperforms ViT$\dagger$ significantly in all four leave-one-out comparisons, indicating the effectiveness of our approach in adapting pre-trained models for domain generalization performance in unseen domains. Additionally, our approach outperforms recent state-of-the-art methods, such as SSAN-R (CVPR2022) \cite{SSAN} and MettaPattern (TIFS2022) \cite{metapattern}, achieving the best results across all four leave-one-out experiments and demonstrating the cross-domain generalization performance of our method. 
%\vspace{-2mm}
\subsubsection{Limited Source Domains} 
%In addition, we assess the performance of our proposed method under the constraint of limited source domains. Adhering to the MICO protocol, we utilize data from only two source domains during the training process. As depicted in Table~\ref{tab:sota-MI-CO}, our method surpasses other state-of-the-art techniques by a considerable margin in terms of HTER and AUC performance. This result substantiates the effectiveness of our method even when fewer source domains are available, further demonstrating its robustness and adaptability.
Furthermore, we evaluate the performance of our proposed method when only a limited number of source domains are available. Modified from the MICO protocol, we use data from only two source domains (``M\&I'') during the training process, and present the results in Table~\ref{tab:sota-MI-CO}. In the experiments of ``M\&I to O'', our method performs comparably with the state-of-the-art AMEL method \cite{zhou2022adaptive} and outperforms AMEL by a clear margin in ``M\&I to C''. Overall, our method can also achieve promising cross-domain performance when the number of source domain data is limited. 
%\vspace{-2mm}
\subsubsection{Few-shot Cross-Domain Evaluation}
The domain generalization evaluation previously mentioned was conducted in a zero-shot cross-domain setting, wherein no target domain data was available during the training phase. In practical applications, however, it is plausible that a small amount of target domain data may be collected to adapt a model to a new target domain. Therefore, we also assessed our method in a few-shot cross-domain setting. Following \cite{eccv2022vit}, we performed 5-shot cross-domain experiments and displayed the results in Table~\ref{tab:5-shot}. To ensure a fair comparison, the results of ``ViTAF$\dagger$'' \cite{eccv2022vit} excluded the CelebA-Spoof dataset \cite{CelebA-Spoof} as supplementary training data \cite{eccv2022vit}, which is consistent with our ViT-S-Adapter. Our S-Adapter demonstrated increased efficiency in 5-shot cross-domain testing, yielding superior results compared to the linear-layer-based adapter employed in \cite{eccv2022vit}. Consequently, our S-Adapter remains promising in the few-shot cross-domain setting.
%\vspace{-3mm}
\subsection{Unseen Attack Detection Evaluation}
While the above cross-domain evaluation only involves 2D attack examples (\textit{i.e.} print, replay), we are aware that there are more attack types appearing, such as mask attacks \cite{SiW-M-2019}. Therefore, it is necessary to evaluate the proposed method's performance in detecting various types of attacks, especially attack types unseen in the training. As such, we utilize the unseen attack evaluation protocol based on the SiW-M dataset \cite{SiW-M-2019}. 
As listed in Table~\ref{tab:intra_SiWM}, the SiW-M dataset includes replay attacks, print attacks, mask attacks, makeup attacks, and partial attacks. 
The leave-one-out unseen attack protocol in the SiW-M dataset \cite{SiW-M-2019} leaves one attack type as the unseen target attack type in the testing. Overall, we can see our method achieves the best ACER results in detecting unseen attack types and surpasses the state-of-the-art method by a significant margin ($>$ 5\%) in terms of ACER and EER. Therefore, our S-Adapter is still effective in detecting out-of-distribution attack types.

\begin{table*}[htbp]
  \centering
  \caption{Results of our S-Adapter and TSR for different ViT backbones: ViT-Large, ViT-Small, and ViT-Tiny. }
    \begin{tabular}{|c|l|c|c|c|c|c|c|c|c|}
    \hline
    \multirow{2}[4]{*}{Backbone} & \multicolumn{1}{c|}{\multirow{2}[4]{*}{Adapter}} & \multicolumn{2}{c|}{C\&I\&O to M} & \multicolumn{2}{c|}{O\&M\&I to C} & \multicolumn{2}{c|}{C\&M\&O to I} & \multicolumn{2}{c|}{C\&I\&M to O} \bigstrut\\
\cline{3-10}          &       & HTER (\%)  & AUC (\%)   & HTER (\%)  & AUC (\%)   & HTER (\%)  & AUC (\%)  & HTER (\%)  & AUC (\%) \bigstrut\\
    \hline
    \multirow{3}[6]{*}{ViT-Tiny} & S-Adapter w/o hist ($\theta=0$) & 13.04  & 94.17  & 20.68  & 87.90  & 27.08  & 71.43  & 21.28  & 87.13  \bigstrut\\
\cline{2-10}          & S-Adapter & 14.95  & 91.89  & 17.75  & 90.68  & 23.46  & 75.94  & 21.38  & 86.91  \bigstrut\\
\cline{2-10}          & S-Adapter-TSR & \textbf{9.92 } & \textbf{96.00 } & \textbf{16.00 } & \textbf{92.22 } & \textbf{19.41 } & \textbf{83.67 } & \textbf{16.24 } & \textbf{90.91} \bigstrut\\
    \hline
    \multirow{3}[6]{*}{ViT-Small} & S-Adapter w/o hist ($\theta=0$) & 11.40  & 95.64  & 17.32  & 91.20  & 19.05  & 89.17  & 15.88  & 91.04  \bigstrut\\
\cline{2-10}          & S-Adapter & 10.52  & 95.66  & 12.11  & 94.06  & 20.02  & 89.85  & 12.74  & 93.10  \bigstrut\\
\cline{2-10}          & S-Adapter-TSR & \textbf{9.23 } & \textbf{96.56 } & \textbf{11.55 } & \textbf{94.90 } & \textbf{12.79 } & \textbf{93.15 } & \textbf{12.32 } & \textbf{94.84 } \bigstrut\\
    \hline
    \multirow{3}[6]{*}{ViT-Large} & S-Adapter w/o hist ($\theta=0$) & 11.04  & 95.56 & 8.03  & 97.27  & 13.74  & 86.62  & 11.13  & 94.93  \bigstrut\\
\cline{2-10}          & S-Adapter & 4.04  & 99.09  & 7.57  & 96.86  & 10.33  & 96.06  & 11.53  & 95.09  \bigstrut\\
\cline{2-10}          & S-Adapter-TSR & \textbf{2.90 } & \textbf{99.48 } & \textbf{7.34 } & \textbf{97.63 } & \textbf{8.54 } & \textbf{97.17 } & \textbf{8.20 } & \textbf{97.69 } \bigstrut\\
    \hline
    \end{tabular}%
  \label{tab:different_backbone}%
\end{table*}%

\subsection{Ablation Study}
We conduct the ablation study to investigate the effectiveness of our proposed S-Adapter and the TSR under different $\lambda$.

\subsubsection{Effectiveness of S-Adapter}  
\begin{figure}[t]
    \centering
    \includegraphics[width=0.95\linewidth]{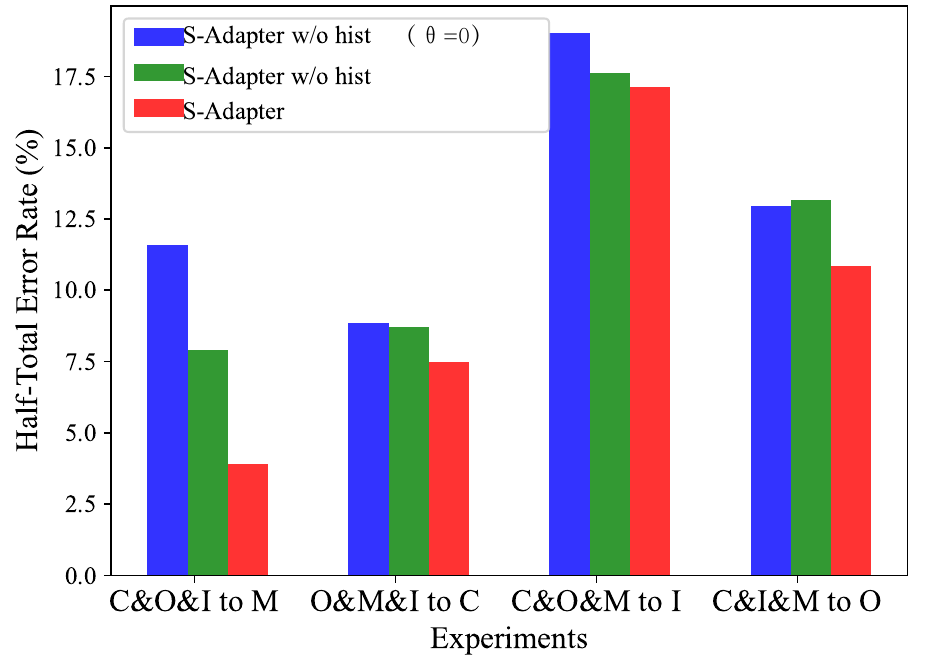} 
    %\vspace{-5mm}
    \caption{Ablation study about adapters. Red bars convey the results of our S-Adapters. Green bars convey the results of removing the token histogram from our S-Adapters. Blue bars convey the results of further removing the gradient information ($\theta=0$).}\label{fig:adapters}
    %\vspace{-6mm}
    \end{figure}

In this study, we investigate the effectiveness of our S-Adapter in a series of cross-domain experiments without the TSR. To assess the impact of the histogram layers and token gradient, we conduct experiments under two additional configurations: 1) ``S-Adapter w/o hist'', where histogram layers are removed, and 2) ``S-Adapter w/o Hist($\theta=0$)'', where both histogram layers and token gradient ($\theta=0$) are removed. 
The experimental results are provided in Fig.~\ref{fig:adapters}. It can be seen that our S-Adapter generally outperforms the other two configurations, illustrating the advantages of extracting the token histogram. We observe that the token gradient also contributes to lower HTER values in most cases. However, in the ``C\&I\&M to O'' experiment, the inclusion of token gradient information results in an increased HTER.
We conjecture that this unexpected result may be attributed to the disparity in texture between the low-resolution source domains (I, C, and M) and the high-resolution target domain (O). Although the fine-grained texture information is extracted in the gradient, the domain gap might cause the texture to differ significantly between the low-resolution and high-resolution domains. In contrast, our histogram layers provide a more comprehensive representation of texture information across resolutions. This is evident in the lower HTER achieved by our S-Adapter compared to the other two configurations in the ``C\&I\&M to O'' experiment.
In summary, our proposed S-Adapter demonstrates performance improvements by leveraging statistical information to enhance cross-domain performance, highlighting the benefits of incorporating a token histogram from the token map with the gradient information.

\begin{figure}
    \centering
    \includegraphics[width=0.95\linewidth]{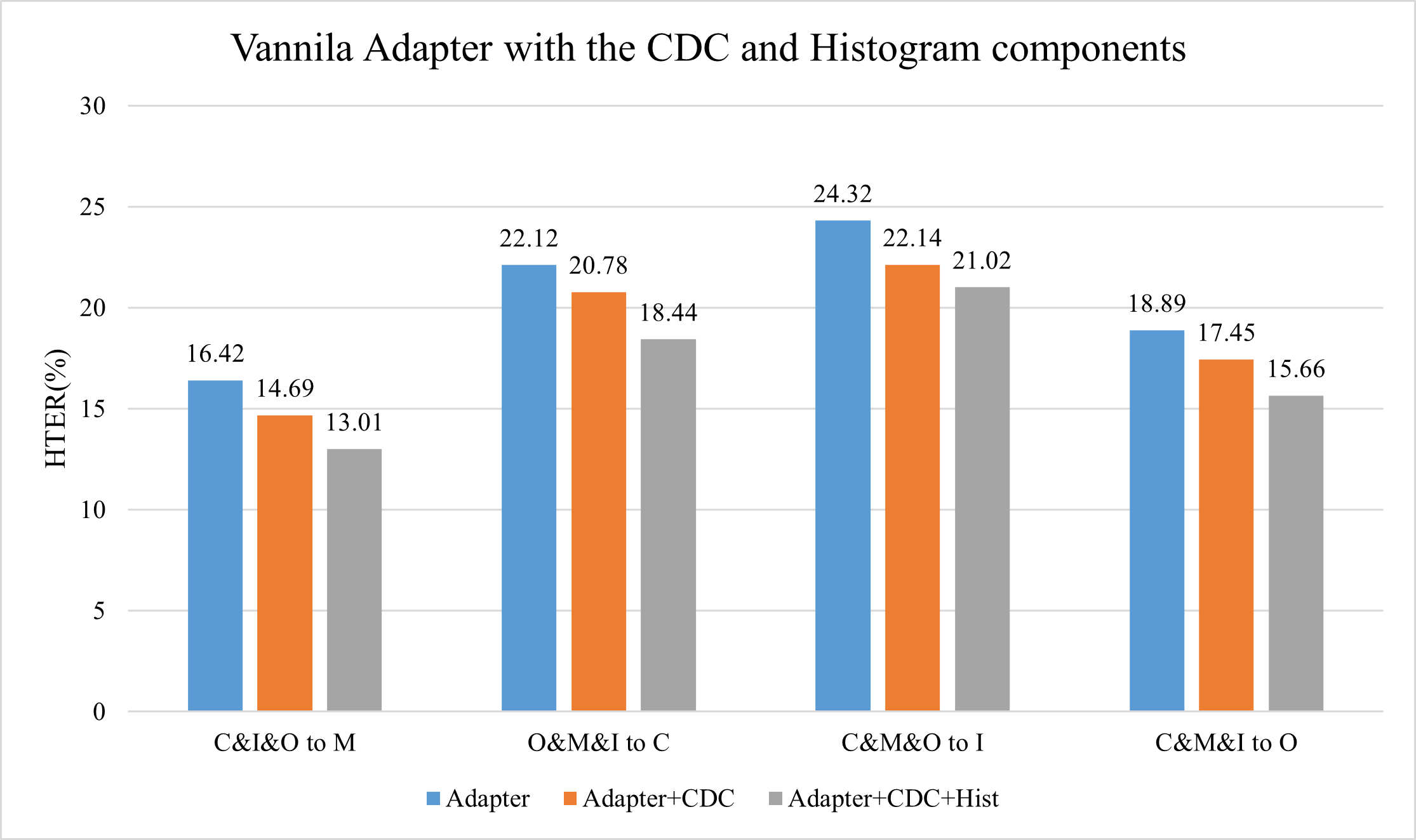}
    \caption{\textcolor{black}{Illustration of the effectiveness of applying our histogram and TSR on a standard vanilla adapter. `Adapter' denotes the standard vanilla adapter. `Adapter+CDC' denotes that a CDC layer is applied after the `Adapter'. `Adapter+CDC+Hist' denotes the Histogram applied after the `Adapter+CDC+Hist'.}}
    \label{fig:linear_with_hist}
\end{figure}

\textcolor{black}{Moreover, we validate the proposed components of the CDC layer and the histogram (Hist) layer with a standard vanilla adapter. To achieve this, after a standard vanilla linear adapter, we add the CDC layer ($\theta=0.7$) and the histogram layer, and the results are denoted and reported in Fig.~\ref{fig:linear_with_hist} as `Adapter+CDC' and `Adapter+CDC+Hist'. As illustrated in Fig.~\ref{fig:linear_with_hist}, our proposed histogram layer also benefits the vanilla adapter to further reduce the HTER. }

\subsubsection{Versatility to other backbones.} \textcolor{black}{In the above experiments, we used ViT-Base to align with previous work. In this study, we examine how the proposed method can be transferred to other backbones, including ViT-tiny, ViT-Small, and ViT-Large, and the results are presented in Table~\ref{tab:different_backbone}. As can be seen in Table~\ref{tab:different_backbone}. The proposed S-Adapter and TSR generally benefit different backbones over the different cross-domain testing experiments. Therefore, our proposed methods are transferable.}

\begin{figure}[t]
    \centering
    \includegraphics[width=0.95\linewidth]{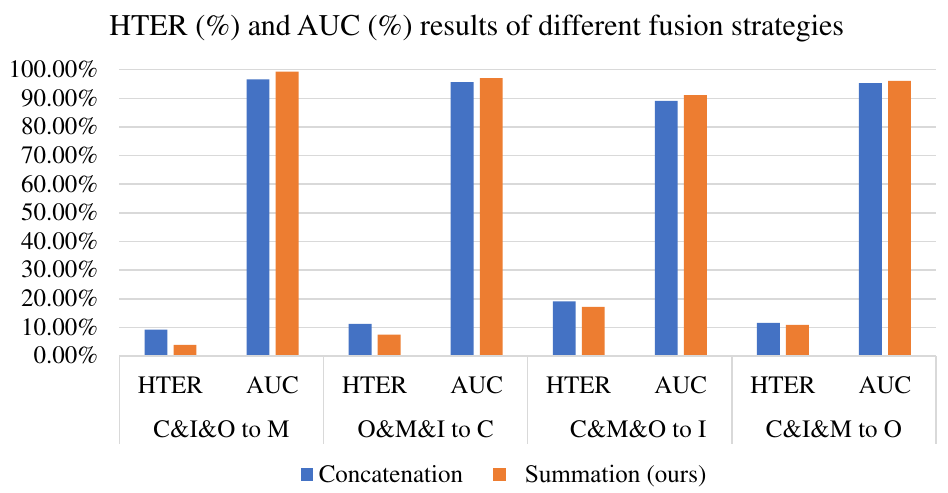} 
    %\vspace{-5mm}
    \caption{Comparison between the fusion strategy of summation and concatenation. Results are in terms of HTER (\%) $\downarrow$ and AUC (\%) $\uparrow$.}\label{fig:fusion}
    %\vspace{-6mm}
\end{figure}

%\textbf{Type of histogram layer}
%We compare two types of histogram layers, Linear Histogram layers and RBF layers to study the difference.

\begin{figure}[t]
    \centering
    \includegraphics[width=0.95\linewidth]{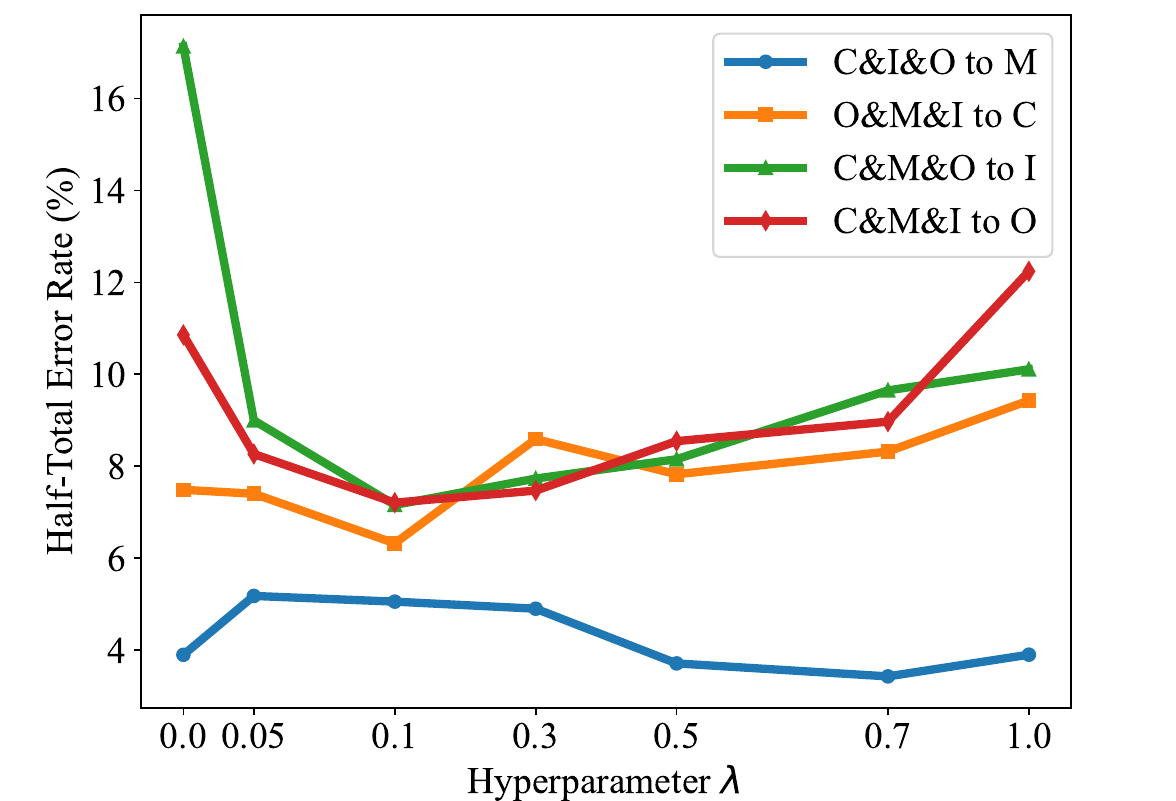}
      %\vspace{-2mm}
    \caption{The curve of HTER (\%) over different $\lambda$ on the leave-one-out cross-domain experiments.} \label{fig:lambda}
    %\vspace{-4mm}
\end{figure}

\begin{figure*}[t]
    \centering
    \includegraphics[width=1.0\linewidth]{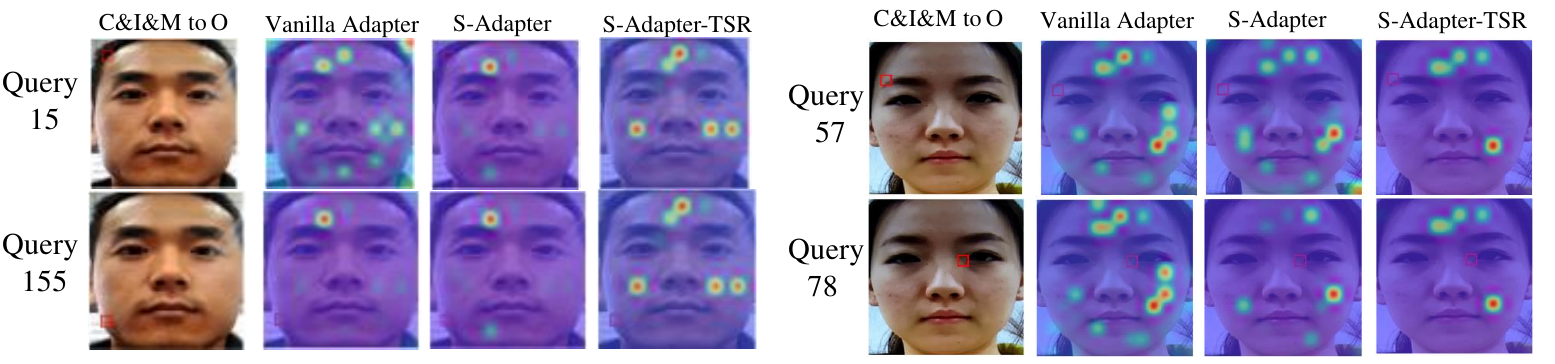} 
    %\vspace{-3mm}
    \caption{Visualization of the maps of self-attention of a query patch. The left part shows an image example's self-attention maps when the 15$^{th}$ (top) and 155$^{th}$ (bottom) are the query patch respectively. The examples are from the OULU-NPU dataset. The vanilla Adapter is based on ViT-S-Adapter but without the statistical tokens (histogram). Red means high attention and blue means low attention. \textcolor{black}{Given that the summation of the attention map within a single image is ``1", the more attention on the background areas, the less attention on the face areas.} Best viewed in color. }\label{fig:vis}
    %\vspace{-4mm}
\end{figure*}

\begin{figure}
    \centering
    \includegraphics[width=0.9\linewidth]{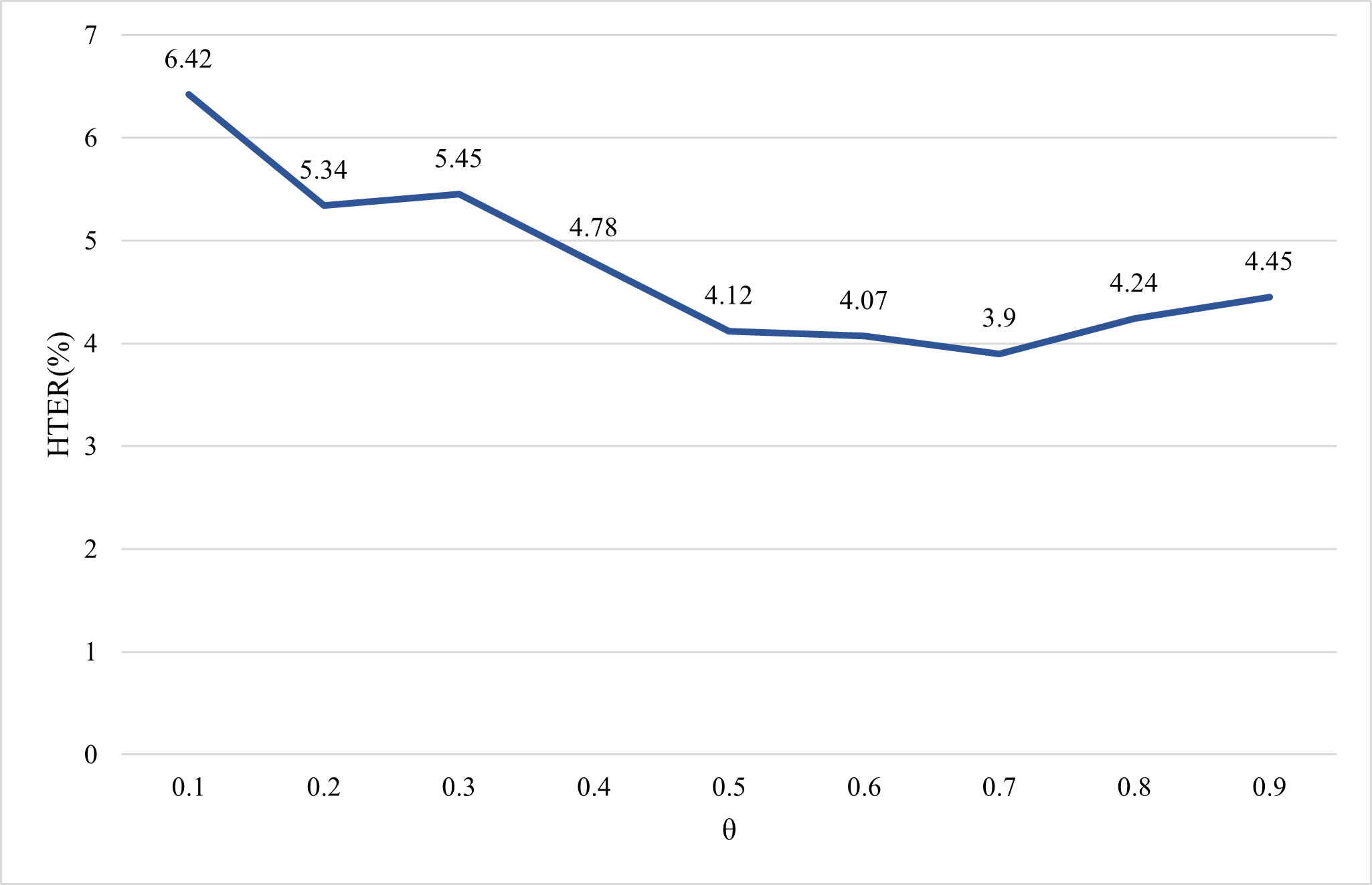}
    \caption{\textcolor{black}{The HTER results reported on the `C\&I\&O to M' experiment with $\theta$ adjusted from 0.1 to 0.9 with our S-Adapter.}}
    \label{fig:hter_theta}
\end{figure}

\subsubsection{Effectiveness of TSR and $\lambda$} 
We investigate the effectiveness of our proposed TSR by altering the balancing ratio $\lambda$, and the experimental results are provided in Fig.~\ref{fig:lambda}. When $\lambda=0$, TSR is not employed in the optimization process, serving as a baseline. Our proposed TSR demonstrates its ability to reduce style variance and facilitate a more generalized model, as evidenced by the further reduction of HTER results when a $\lambda > 0$ is applied across the four experiments. It is important to note that the optimal $\lambda$ varies under different source domain settings, which is reasonable considering that different combinations of source domains possess varying style variations. However, we do not recommend using a large $\lambda$, as the model would overfit to aligning the token style and become less effective in classification. Generally, a small $\lambda$ (\textit{e.g.} 0.1) is advised.
%\vspace{-2mm}
%\subsection{Visual Analysis}

\subsubsection{Study on the fusion strategy} \textcolor{black}{As depicted in Fig.~\ref{fig:framework},  $Y$ is fused with the original token input $X$ via a summation operation. It could be argued that $Y$ is about the statistical information, which is different from $X$ and thus it does not make sense to do the summation between $X$ and $Y$. We point out that it is $Z^{Hist}$ that is about the statistical information, and $Y$ has been transformed and aligned to $X$ in terms of the semantics and dimensions via the ``dimension up" (linear layer), as shown in Fig.~\ref{fig:framework}. Thus, it makes sense to fuse $Y$ with $X$ via the summation. Nevertheless, we conduct experiments by concatenating $Y$ to the summation result between $X$ and the results after the MHSA. After the concatenation, the tokens' dimension is doubled (768$\times$2). To fix the dimension mismatch, we forward the concatenated tokens to a linear layer to reduce the dimension back to 768 to match the next layer's input dimension. The experimental results are shown in Fig.~\ref{fig:fusion}. We can see that the concatenation fusion is providing poorer results. We conjecture that the extra linear layers increase the model complexity and make the model prone to overfitting. }

\subsubsection{The impact of $\theta$} 
\textcolor{black}{
To further study the effectiveness of $\theta$ on the statistic histogram, we conduct experiments to observe how adjusting $\theta$ from 0.1 to 0.9. As can be seen in Fig.~\ref{fig:hter_theta}, when $\theta$ is around 0.5, the model exhibits satisfactory performance ($\approx$ 4\% HTER), as the component from vanilla convolutions and its gradients are well balanced. When $\theta=0.7$, the lowest HTER is achieved, which aligns with the recommendation by \cite{CDCN-CVPR-2020}.
}

\subsection{Attention Map Visualization} \textcolor{black}{Since the core of ViT is the self-attention mechanism, we visualize the results of self-attention from the last transformer block to analyze the model behavior. In a self-attention layer, there are $N$ tokens of patches as input. Each output token is a weighted summation of the input tokens, where the weight matrix for each query token is calculated by self-attention. For example,  in Fig.~\ref{fig:vis}, the top left image shows that the 15$^{th}$ patch (in red rectangle) is the query, and the right sides are the attentions of the other patches, a.k.a, the weighted matrix for the summation. A patch with red attention means it contributes significantly to the output tokens of the query patch.  The first row shows the attention when the 15$^{th}$ patch is the query patch, and the second row shows the attention when the 155$^{th}$ patch is the query patch. \textcolor{black}{We can see that our proposed S-Adapter and TSR can help the model to focus more consistently and intensively on the face areas (more redness than the background area),  which is compared within a single image, rather than between multiple images.} For example, we observe that our ViT-S-Adapter-TSR is consistent in the attention maps given different queries. The output attention maps of Query 15 and Query 155 are similar, meaning that the model is ``keeping focus'' on a similar area. On the other hand, the ViT with vanilla adapter is paying attention to various areas, since the attention maps of the two queries (15$^{th}$ and 155$^{th}$ patch) vary hugely. When the query is the 15$^{th}$ patch, the ViT-Adapter even focuses on the background region, instead of mining spoofing cues from the face regions. Our observations are aligned with the \cite{pan2022less} that a vision transformer paying less attention can achieve more generalization. The observations are also consistent in the examples in the right part of Fig.~\ref{fig:vis}}.

\subsection{Visualization of t-SNE } 
\begin{figure}
    \centering
    \includegraphics[width=\linewidth]{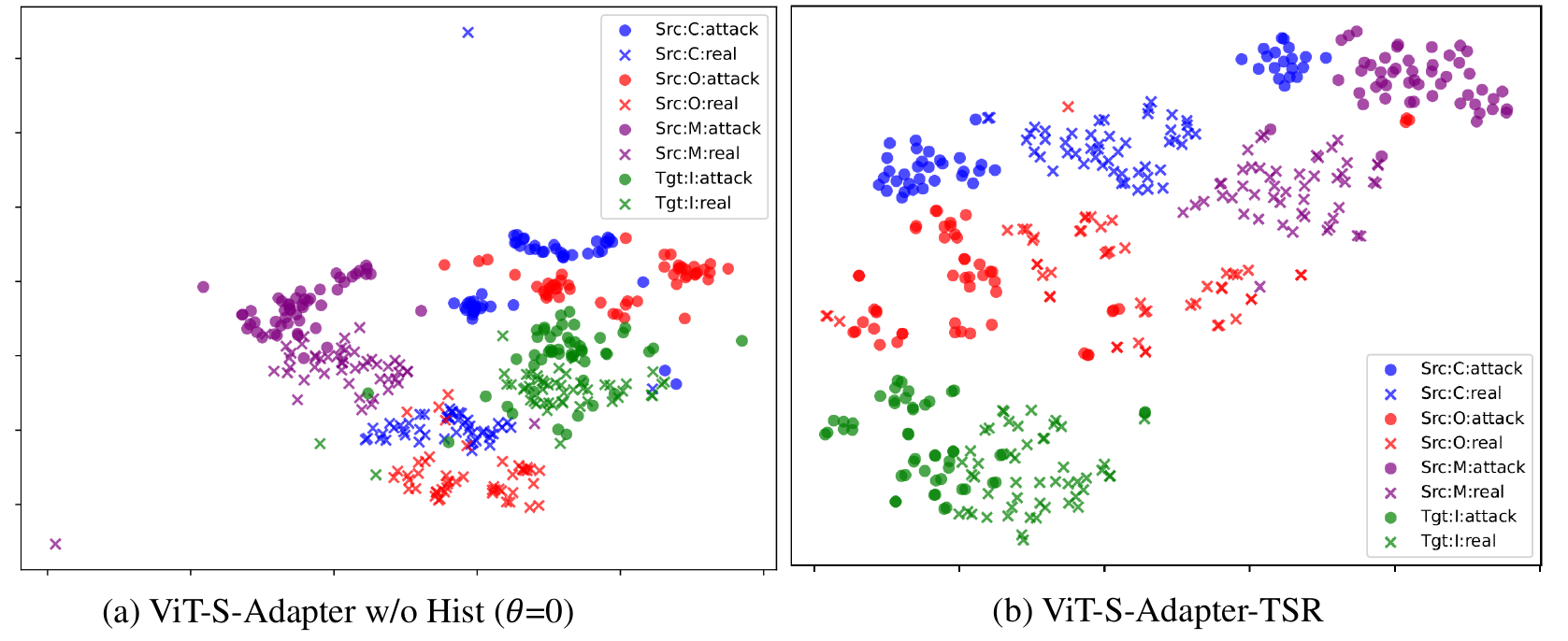}
    %\vspace{-8mm}
    \caption{The visualization of t-SNE conducted on the class token of two different adapters in the experiments of `C\&O\&M to I'. (a) is the ViT-S-Adapter without the statistical and fine-grained information ($\theta=0$), which serves as the baseline. (b) represents our ViT-S-Adapter with the TSR.}\label{fig:tsne}
    %\Description{figure description}
\end{figure}

\textcolor{black}{We also use the t-SNE technique to visualize the distribution of class tokens (features) from the last transformer block to show the effectiveness of the proposed method. In Fig.~\ref{fig:tsne}, the models are trained by the combined dataset `C\&O\&M' and the target test data is the `I'. Fig.~\ref{fig:tsne}(a) represents the t-SNE visualization of the ViT-S-Adapter without the statistical (histogram) and fine-grained information, which serves as the baseline. Fig.~\ref{fig:tsne}(b) represents the result of ViT-S-Adapter optimized with TSR. As can be seen in Fig.~\ref{fig:tsne}, the `real' and ``attack'' samples from the target (tgt) domain are overlapped and cannot be well separated (green color). By contrast, our proposed ViT-S-Adapter-TSR has better generalization capability that the `real' and `attack' samples from the target domains are well separated.}
%\vspace{-2mm}
\subsection{Overhead Analysis}
We analyzed the overhead introduced by the S-Adapter, which contains additional parameters, by utilizing an open-source library\footnote{https://github.com/Lyken17/pytorch-OpCounter} to collect data on Multiply-Accumulate operations (MACs) and the number of parameters for both the ViT (Base) model and the ViT with our S-Adapter. As illustrated in Table~\ref{tab:macs}, the inclusion of the S-Adapter results in a modest increase of only 0.45\% in MACs and a mere 0.38\% in parameters overhead. Given the substantial improvement in zero/few-shot cross-domain performance and unseen attack detection capabilities provided by our S-Adapter, this minimal overhead is well justified.

\begin{table}[t]
  \centering
  \caption{Inference time overhead analysis. The MACs (G) indicate the total calculation operations and the Params (M) indicate the amount of parameter and storage needed.}  \label{tab:macs}%
  %\vspace{-2mm}
    \begin{tabular}{|c|c|c|}
    \hline
    Model & MACs (G) & Params (M) \bigstrut\\
    \hline
    ViT   & 33.73 & 85.65 \bigstrut\\
    \hline
    ViT-S-Adapter & 33.88 & 85.98 \bigstrut\\
    \hline
    Increment ($\Delta$)  & 0.15 (0.45\%) & 0.33(0.38\%) \bigstrut\\
    \hline
    \end{tabular}%
%\vspace{-4mm}
\end{table}%

%\vspace{-0mm}
\section{Conclusion and Future Work}
In conclusion, Face Anti-Spoofing is important for the security and integrity of face recognition systems by identifying and thwarting malicious attacks. Although there has been significant progress in recent years, the domain shift problem continues to pose a challenge to a model's cross-domain generalization performance. To tackle this issue, we introduced a novel method, S-Adapter, which effectively utilizes knowledge from pre-trained Vision Transformers (ViT) for cross-domain generalization in face anti-spoofing. Our method of S-Adapter employs the histogram information of transformer tokens and incorporates our proposed Token Style Regularization (TSR) to learn more domain-invariant feature representations. Our comprehensive experiments reveal that the proposed S-Adapter and TSR method surpasses existing techniques, achieving state-of-the-art performance on several face anti-spoofing benchmarks of zero/few-shot cross-domain evaluations, and unseen attack detection. 

In the future, the proposed S-Adapter may be extended to other similar tasks such as inpainting detection, deepfake detection, and recaptured document detection problems as well. Besides, how the token histogram can be used with LoRA and Prompt can be explored in the future.

% TODO:
% Explain why not better than the AMEL
% Discuss more about the fusion strategy, justification
% Revise the figure: Average -> Summation
% 
\bibliographystyle{ieeetr}
\bibliography{sample-base}
\begin{IEEEbiography}[{\includegraphics[width=1in,height=1.25in,clip,keepaspectratio]{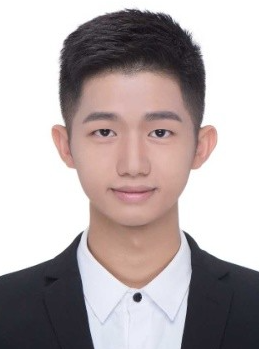}}]
	{Rizhao Cai} received his B.Eng degree in electronic information engineering from  Shenzhen University, China in  2018. After that, he worked as a project officer in the Rapid-Rich Object Search (ROSE) Lab and NTU-PKU  Joint  Research  Institute, leading computer vision projects collaborated with industrial partners. Since 2020, he has been a Ph.D candidate at Nanyang  Technological  University,  Singapore. His research interests include computer vision and biometric/AI security.
\end{IEEEbiography}

\begin{IEEEbiography}[{\includegraphics[width=1.05in,height=1.3in,clip,keepaspectratio]{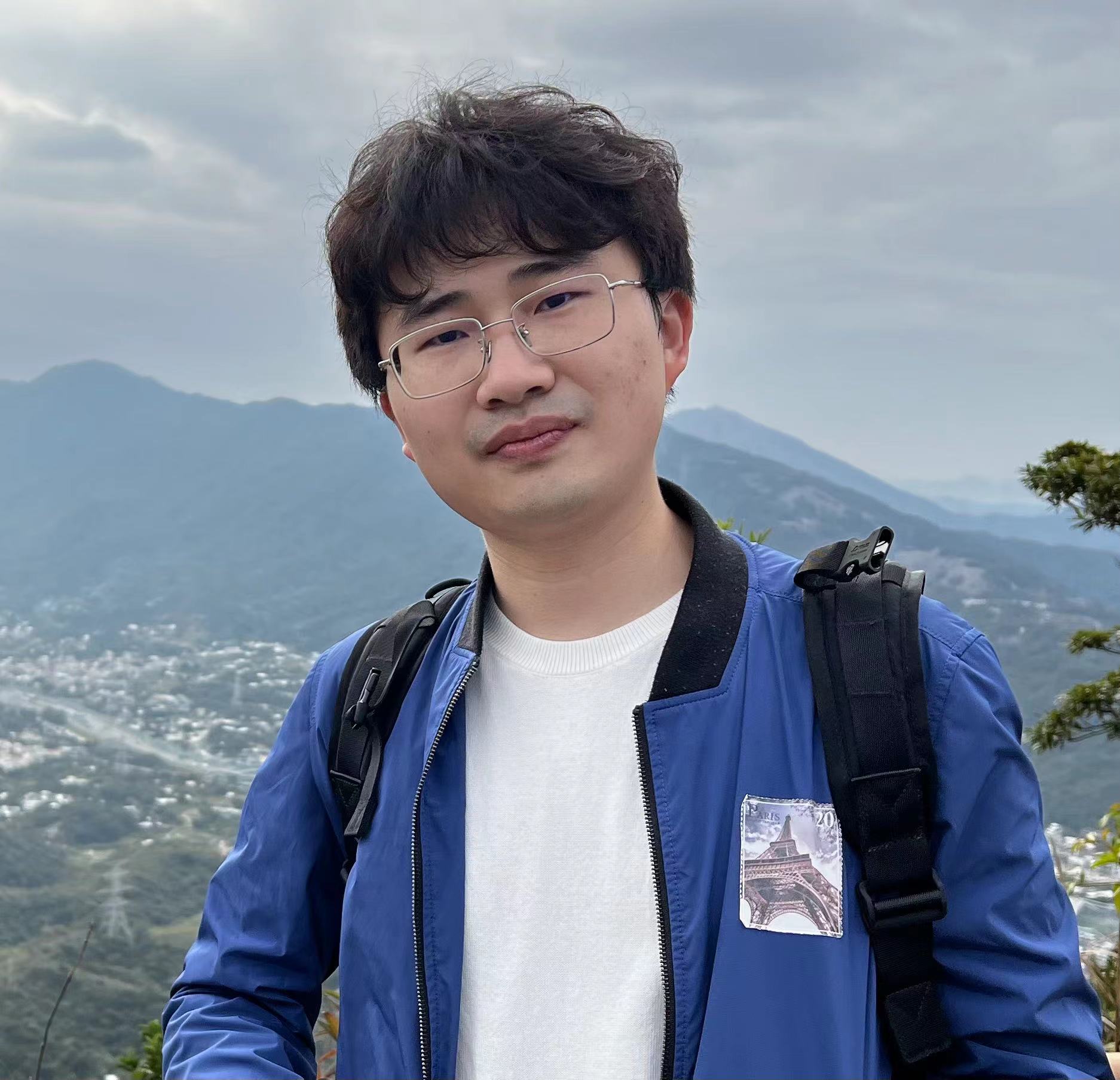}}]{Chenqi Kong} received the B.S. and M.S. degrees from the Harbin Institute of Technology, Harbin, China, in 2017 and 2019, respectively, and the Ph.D. degree from the Department of Computer Science, City University of Hong Kong, Hong Kong, SAR, China, in 2023. Currently, he is a Research Fellow with the School of Electrical and Electronic Engineering, Nanyang Technological University, Singapore. His current research interests include AI security and multimedia forensics. He was a recipient of the National Scholarship, the Gold Medalist at the International Exhibition of Inventions, Geneva, and the Research Tuition Scholarship.
\end{IEEEbiography}

 \begin{IEEEbiography}[{\includegraphics[width=1in,height=1.5in,clip,keepaspectratio]{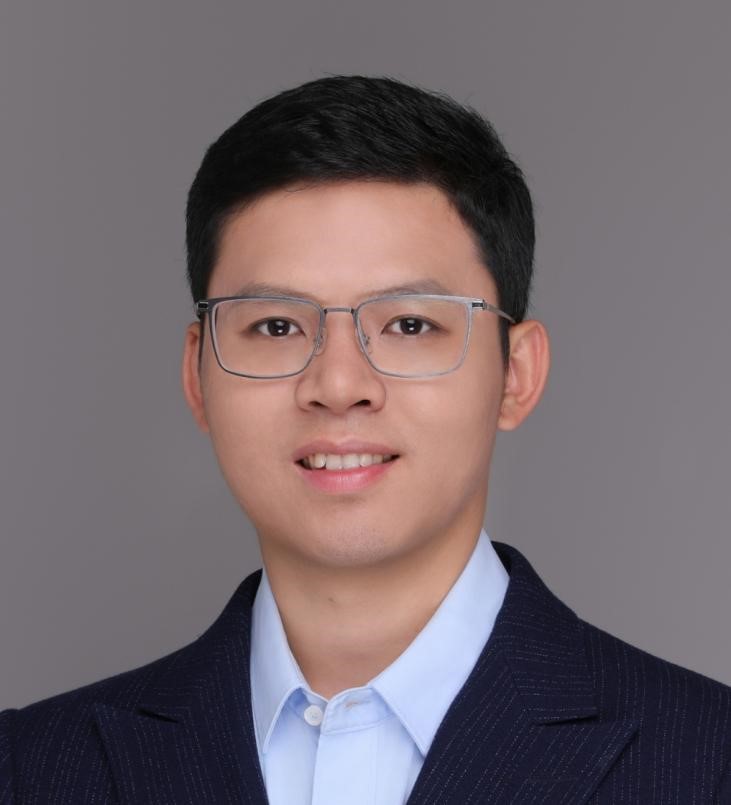}}]
{Zitong Yu} (Senior Member, IEEE) received the Ph.D. degree in Computer Science and Engineering from the University of Oulu, Finland, in 2022. Currently, he is an Assistant Professor at Great Bay University, China. He was a Postdoctoral researcher at ROSE Lab, Nanyang Technological University. He was a visiting scholar at TVG, University of Oxford, from July to November 2021. His research interests include human-centric computer vision and biometric security. He was a recipient of IAPR Best Student Paper Award, IEEE Finland Section Best Student Conference Paper Award 2020, second prize of the IEEE Finland Jt. Chapter SP/CAS Best Paper Award (2022), and World’s Top 2% Scientists 2023 by Stanford. 
\end{IEEEbiography}

 \begin{IEEEbiography}[{\includegraphics[width=1in,height=1.5in,clip,keepaspectratio]{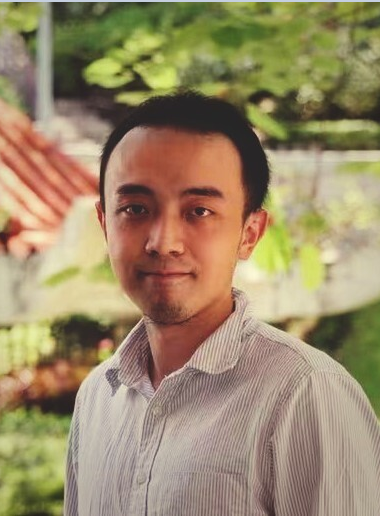}}]
 {Haoliang Li} received the B.S. degree in communication engineering from University of Electronic Science and Technology of China (UESTC) in 2013, and his Ph.D. degree from Nanyang Technological University (NTU), Singapore in 2018. He is currently an assistant professor in Department of Electrical Engineering, City University of Hong Kong. His research mainly focuses on AI security, multimedia forensics and transfer learning. His research works appear in international journals/conferences such as TPAMI, IJCV, TIFS, NeurIPS, CVPR and AAAI. He received the Wallenberg-NTU presidential postdoc fellowship in 2019, doctoral innovation award in 2019, and VCIP best paper award in 2020.  
\end{IEEEbiography}

 \begin{IEEEbiography}[{\includegraphics[width=1in,height=1.5in,clip,keepaspectratio]{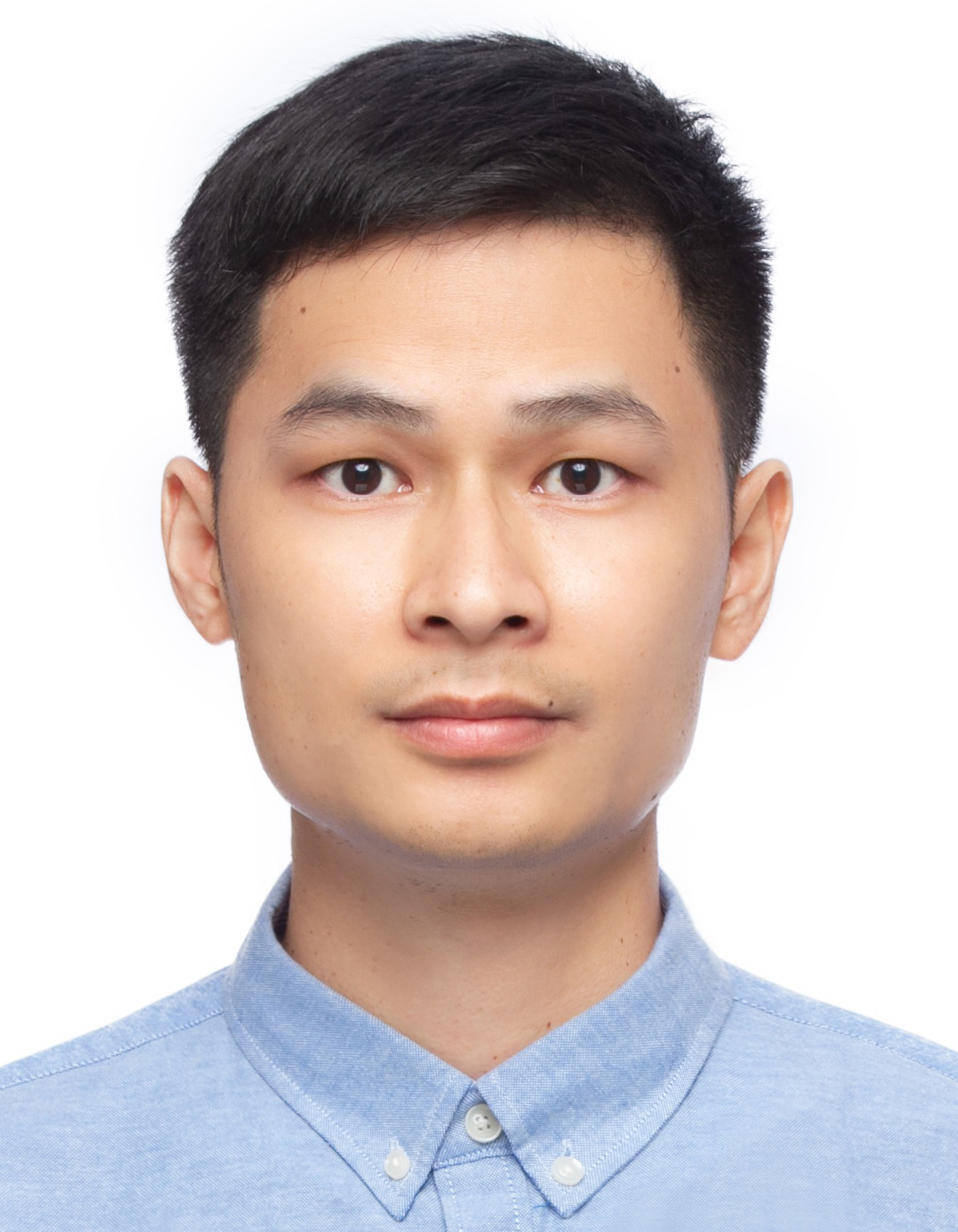}}]
 {Changsheng Chen} (S'09-M'15-SM’20) received the B.Eng. degree in software engineering from Sun Yat-sen University, Guangzhou, China in 2008 and the Ph.D. degree in Electrical and Electronic Engineering from Nanyang Technology University, Singapore in 2013. From 2013 to 2015, he worked as a PostDoc research associate at the HKUST Barcode Group, Department of Electronic and Computer Engineering, Hong Kong University of Science and Technology. Since 2016, he has been with the Guangdong Key Laboratory of Intelligent Information Processing and Key Laboratory of Media Security, College of Electronics and Information Engineering, Shenzhen University, Shenzhen, China, where he is currently an Associate Professor. His current research interests include multimedia forensics and security, pattern recognition and machine learning.
\end{IEEEbiography}

 \begin{IEEEbiography}[{\includegraphics[width=1in,height=1.5in,clip,keepaspectratio]{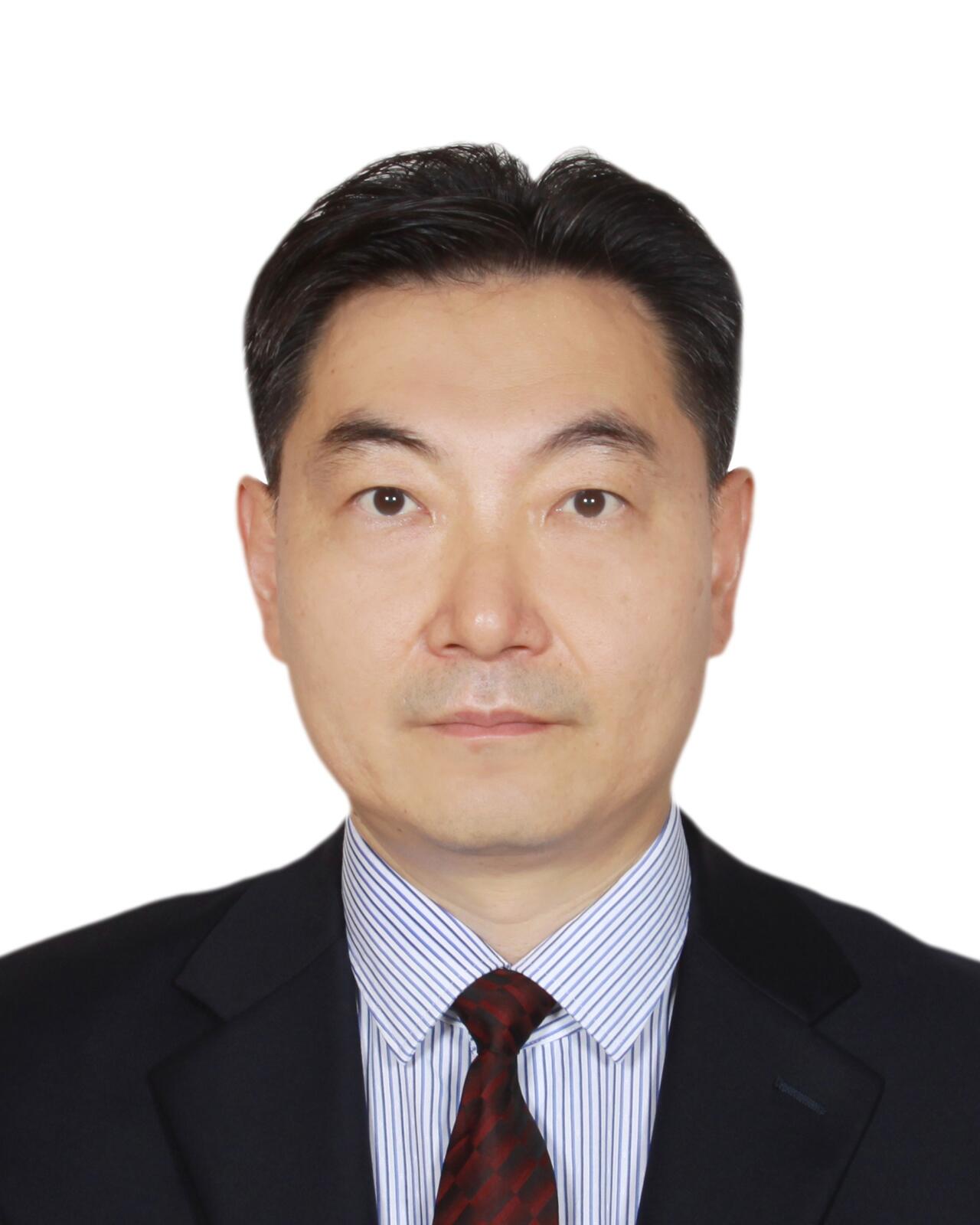}}]
 {Yongjian Hu} received the Ph.D. degree in communication and information systems from South China University of Technology in 2002. Between 2000 and 2004, he visited City University of Hong Kong four times as a researcher. From 2005 to 2006, he worked as Research Professor in SungKyunKwan University, South Korea. From 2006 to 2008, he worked as Research Professor in Korea Advanced Institute of Science and Technology, South Korea. From 2011 to 2013, he worked as Marie Curie Fellow in the University of Warwick, UK. Now He is full Professor with the School of Electronic and Information Engineering, South China University of Technology, China. He is also a research scientist with China-Singapore International Joint Research Institute. Dr. Hu is Senior Member of IEEE and has published more than 130 peer reviewed papers. His research interests include image forensics, information security, and deep learning.
\end{IEEEbiography}

 \begin{IEEEbiography}[{\includegraphics[width=1in,height=1.5in,clip,keepaspectratio]{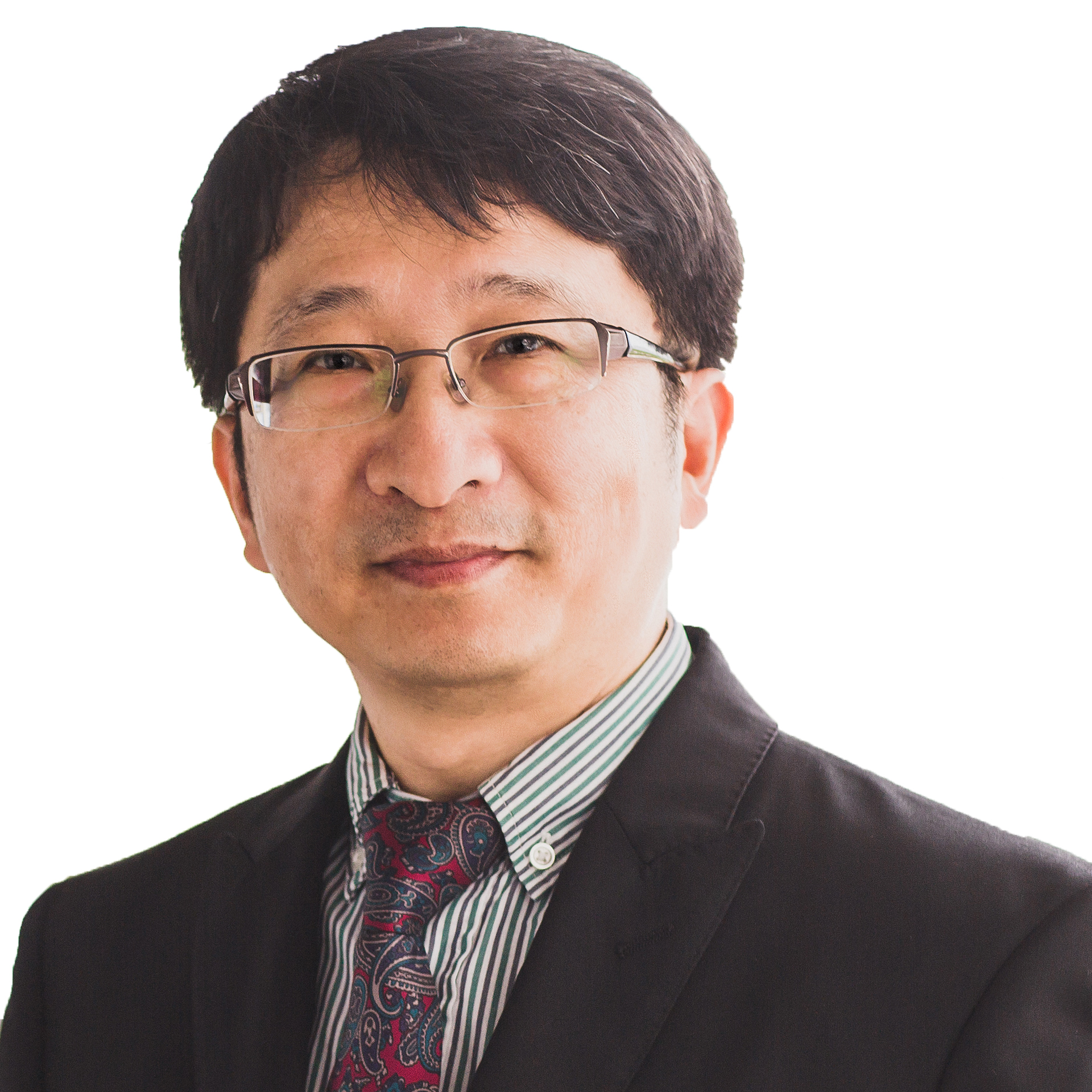}}]
 {Prof. Alex Kot} has been with the Nanyang Technological University, Singapore since 1991. He was Head of the Division of Information Engineering and Vice Dean Research at the School of Electrical and Electronic Engineering. Subsequently, he served as Associate Dean for College of Engineering for eight years. He is currently Professor and Director of Rapid-Rich Object SEarch (ROSE) Lab and NTU-PKU Joint Research Institute. He has published extensively in the areas of signal processing, biometrics, image forensics and security, and computer vision and machine learning. Dr. Kot served as Associate Editor for more than ten journals, mostly for IEEE transactions. He served the IEEE SP Society in various capacities such as the General Co-Chair for the 2004 IEEE International Conference on Image Processing and the Vice-President for the IEEE Signal Processing Society. He received the Best Teacher of the Year Award and is a co-author for several Best Paper Awards including ICPR, IEEE WIFS and IWDW, CVPR Precognition Workshop and VCIP. He was elected as the IEEE Distinguished Lecturer for the Signal Processing Society and the Circuits and Systems Society. He is a Fellow of IEEE, and a Fellow of Academy of Engineering, Singapore. 

\end{IEEEbiography}

\end{document}